\documentclass{article}
\usepackage[utf8]{inputenc}
\usepackage{authblk}
\usepackage{setspace}
\usepackage[margin=1.25in]{geometry}
\usepackage{graphicx}
\usepackage{subcaption}
\usepackage{amsmath}
\usepackage{lineno}
\usepackage{tabularx}
\usepackage{gensymb}
\usepackage[square,numbers]{natbib}
\usepackage{url}
\usepackage{amssymb}

\title{A text-based, generative deep learning model for soil reflectance spectrum simulation in the VIS-NIR (400-2499 nm) bands}

\author[1*]{Tong Lei}
\author[1]{Brian N. Bailey}

\affil[1]{Department of Plant Sciences, University of California, Davis, CA, USA}

\affil[*]{tlei@ucdavis.edu}

\date{}

\begin{document}

\maketitle

\begin{abstract}

Soil spectral reflectance is a necessary input for land surface and radiative transfer models, and can be used to infer soil properties. Numerous models have been developed based on mechanistic approaches, each with their own limitations. Mechanistic models based on radiative transfer theory are usually based on only a few input soil properties, whereas data-drive approaches are limited by high non-uniformity of available published datasets that severely limits the amount of data usable for model calibration. To address these limitations, a fully data-driven soil optics generative model (SOGM) for simulation of soil reflectance spectra based on soil property inputs was developed based on the denoising diffusion probabilistic model (DDPM). The model was trained on an extensive dataset comprising nearly 180,000 soil spectra-property set pairs from 17 published datasets. The model generates soil reflectance spectra from text-based inputs describing soil properties and their values rather than only numerical values and labels in binary vector format, which means the model can handle variable formats for property reporting. Because the model is generative, it can simulate reasonable output spectra based on an incomplete set of available input properties, which becomes more constrained as the input property set becomes more complete. Two additional sub-models were also built to complement the SOGM: a spectral padding model that can fill in the gaps for spectra shorter than the full visible-near-infrared range (VIS-NIR; 400 to 2499 nm), and a wet soil spectra model that can estimate the effects of water content on soil reflectance spectra given the dry spectrum predicted by the SOGM. The SOGM was up-scaled by coupling with the Helios 3D plant modeling software, which allowed for generation of synthetic aerial images of simulated soil and plant scenes. It can also be easily integrated with other soil-plant radiation models used for remote sensing research such as PROSAIL. The testing results of the SOGM on new datasets that not included in model training demonstrated that the model can generate reasonable soil reflectance spectra based on available property inputs. Results also show soil clay/sand/silt fraction, organic carbon content, nitrogen content, and iron content tended to be important properties for spectra simulation. Inclusion of some trace minerals like nickel as model inputs decreased model performance because of their low concentrations and large propensity for ground-truth measurement error. The presented models are openly accessible on: \url{https://github.com/GEMINI-Breeding/SOGM_soil_spectra_simulation}.

\end{abstract}

\section{Introduction}\label{intro}

The spectral reflectance of soil determines how radiation of varying wavelength interacts with soil constituents, and thus indirectly contains information about the state and composition of the soil. It is therefore commonly used as a remotely measurable quantity for inferring information about the soil \citep{sadeghi2018particle,yuan2020wavelength}, or as an input for models seeking to predict how radiation interacts with the soil surface. Land surface models are widely used to predict fluxes of energy, mass, and momentum at Earth's surface, usually across large scales, and require soil reflectance spectra as an input to predict surface radiative energy exchange. Examples include 4SAIL2 \citep{verhoef2007coupled}, PROSAIL \citep{prosail}, soil-canopy observation photosynthesis and energy fluxes (SCOPE) model \citep{van2016model}, and soil-plant-atmosphere radiative transfer (SPART) model for satellite measurements \citep{yang2020spart}. In other applications, soil reflectance spectra may be needed as an input for algorithms that use radiative transfer model inversion or machine learning models to determine physical properties of overlying vegetative canopies from remotely sensed images that contain both plant elements and background soil \citep{chen2022unsupervised,chen2022integrating,upreti2019comparison}. 

Because of the scale at which land surface models are often applied (i.e., regional to global), it is not feasible to manually measure the spatial variability in soil reflectance spectra, and thus it must be modeled. Soil reflectance spectra is commonly predicted using soil radiative transfer models that describe the relationship between soil reflectance and soil properties \citep{sadeghi2018particle,marmit,marmit2,wu2023semi,bach1994modelling,yuan2019semi}. The soil reflectance spectra depends on complex relationships between soil structure, aggregate constituents, and water content \citep{wu2023novel}, and thus some input parameters for these models can be difficult to measure at scale, such as the absorption coefficient of soil \citep{yuan2019semi,wu2023novel} or parameters fitted from customized empirical functions \citep{marmit,marmit2}. As a result, these models typically rely on empirical relationships obtained from laboratory samples, and may thus have issues when applied at the field scale. Furthermore, previous soil radiative transfer models have primarily focused on a limited number of factors, such as moisture, organic matter, and particle size \citep{sadeghi2018particle,marmit,marmit2,wu2023semi,bach1994modelling,yuan2019semi}. However, many important soil properties like organic carbon, total nitrogen, and pH values, which also impact soil reflectance spectra, have not been integrated into most previous radiative transfer models for spectral simulation.

The extensive advancement in techniques for soil monitoring has motivated the curation of vast amounts of soil spectra and property ground truth data by various laboratories and institutes around the world \citep{lucas,BSSL,lucas2015,ossl}. Given the limitations of semi-empirical radiative models for soil spectral simulation, a potential alternative is to use a fully data-driven approach that has the ability to incorporate large amounts of existing soil spectra-property datasets. However, the diversity and inconsistency of existing datasets presents additional challenges when using a data-driven approach. Spectra obtained from different instruments may have different measurement wavelengths, and may introduce biases into the data due to variability in the calibration or internal components of each instrument. In previous studies and datasets such as OSSL \citep{ossl}, one approach is to partially crop the data to fit to a wavelength range that is common across all datasets, but this discards a large amount of potentially informative data. \citep{lei2022achieving} developed a transformer-based predictive model that can handle soil spectra obtained by different instruments. However, for soil spectra simulation used in radiative transfer modeling \citep{ganapol1999lcm2,prosail}, where spectra that span the full visible to near-infrared wavelength range are normally required, it becomes necessary to generate the missing wavebands to ensure the completeness of the simulated data.

Another considerable challenge in using a data-driven approach for soil reflectance spectral modeling is that ground truth measurements of soil properties differ substantially across datasets, both in terms of which measurements were collected and how their values are reported. Soil properties are commonly incorporated into models as either quantitative values or qualitative labels \citep{marmit,sadeghi2018particle,lei2022achieving}. However, some soil properties, like total phosphorus content and exchangeable phosphorus content, share complex relationships that are not easily captured using physical models. The same property may be described using different units, such as \% or g/dm$^3$, making direct conversion between them impossible. Additionally, certain property values are represented as text-based labels, such as qualitative land use and land cover classifications, further complicating the integration of these properties into parameterized models. Therefore, directly inputting property values into a parameterized model can lead to several issues: it risks losing valuable connections among different soil properties; handling a combination of labels and numerical values within the same model becomes a challenging task; because different soil properties might be presented, relying solely on the sequence of values to identify specific properties within the model is likely to be an unreliable approach. Consequently, directly inputting text-based descriptions of soil properties into the model might be a better choice, as text descriptions inherently encode the property differences.

In this work, a soil optics generative model (SOGM) for simulation of soil reflectance spectra based on soil property inputs was developed. The modeling approach has three primary novel aspects that distinguish it from previous approaches. First, the model is fully data-driven and is trained on an extensive dataset comprising nearly 180,000 soil spectra and their measured physical properties. Second, it generates soil reflectance spectra from text-based inputs describing soil properties and their values rather than only numerical values and labels in binary vector format. Third, the model is generative, which means that it can simulate output spectra based on an incomplete set of input properties. In contrast to a predictive model which yields identical outputs for identical inputs, a generative model enables varied outputs for identical (incomplete) inputs. This is an important feature for soil optical modeling since many soil properties that can influence the soil reflectance spectra may not explicitly listed in available datasets used for model training, or the user may not know all soil properties when the model is applied for prediction. SOGM is based on the denoising diffusion model \citep{ddpm}, which is popular in deep-learning-based image generation \citep{latentdiffusion}.  Two additional sub-models were created to complement the SOGM: a spectral padding model that can fill in the gaps for spectra shorter than the full visible-near-infrared range (VIS-NIR; 400 to 2499 nm), and a wet soil spectra model that can estimate the effects of water content on soil reflectance spectra given the dry spectrum predicted by the SOGM. 
The SOGM was up-scaled by coupling with the Helios 3D plant modeling software \citep{HELIOS}, which allowed for generation of synthetic aerial images of simulated soil and plant scenes.

\section{Datasets and models}\label{S:model}

\subsection{Datasets}\label{S:data}

In the current study, 17 dry soil VIS-NIR or NIR spectral datasets were utilized for training the spectra padding model and SOGM (Table \ref{tab:data4train}). In certain datasets, such as OSSLJ and LUCAS, spectra obtained from repeated scans on the same sample were retained to ensure a sufficiently large training set, resulting in a total of 177,579 soil spectra. The noisy segments and portions beyond the 400 to 2499 nm range of the spectra were manually excluded. For each spectrum, the first wavelength and the last wavelength, plus 1, that are not divisible by 50 were also omitted (the reason is given in Sect. \ref{S:model:pad}). For example, if the first measured wavelength in the spectrum was 380 nm, it would be increased to 400 nm (wavelengths from 380-399 nm were removed), and for the last wavelength, 2360 nm would be trimmed to 2349 nm. 

The final wavelength ranges for the datasets considered in this study are presented in Table \ref{tab:data4train}. We believe this to be the largest compiled soil VIS-NIR spectra model to date. These datasets were acquired using spectrometers from various manufacturers, including FOSS NIRSystems Inc. (Hilleroed, Denmark), Analytical Spectral Devices (ASD) Inc., (Boulder, Colorado, USA), Bruker Optics (Ettlingen, Germany), Spectral Evolution, Inc. (Haverhill, Massachusetts, USA), Si-Ware Systems (Menlo Park, California, USA), and others. Instrument manufacturer was included as a variable in the model to account for potential biases due to instrument. While a single manufacturer like ASD offers different models such as the FieldSpec3, FieldSpec4, and LabSpec, the differences between versions from the same manufacturer are minor. Consequently, version information was omitted from the model to improve its generalizability (Table \ref{tab:prop}).

Table \ref{tab:data4test} lists three datasets used for testing the spectral padding and diffusion model, comprising a total of 652 dry soil spectra. The original MARMIT2020 data repository \citep{marmit,marmit2} includes 8 sub-datasets \citep{marmit,marmit2,Humper_2015,lesaignoux2013,liu2002,lobell2002moisture,Philpot2014} with a total of 1894 spectra, of which 211 are dry soil spectra. For all 3 datasets, outlier spectra were excluded, such as those with many values higher than 1 or with high noise level, and spectra with no valid properties. ``Valid properties" refers to properties that appeared in the training datasets, including Clay, Silt, Sand, Soil organic matter, CaCO$_3$, Mg, Al, P, Ca, Mn, Fe, Zn, and Ni contents, pH value, Cation-exchange capacity, among others. Some property values were not provided in the original datasets, hence the average valid soil property values for some datasets are not integers in Table \ref{tab:data4test}.

For the wet spectra model, both wet and dry spectra from the MARMIT2020 data repository \citep{marmit,marmit2} and an additional 78 wet and dry spectra from \citep{tian2021soil} were used. More details are provided in Sect. \ref{S:water}.

\begin{table}
\centering
\caption{Datasets for model training}
\begin{tabular}{>{\raggedright\arraybackslash}p{2cm} >{\raggedright\arraybackslash}p{2cm}>{\raggedright\arraybackslash}p{3cm}>{\raggedright\arraybackslash}p{5cm}}
  \hline
  Datasets & Number of spectra & Selected wavelength range & Reference \\
  \hline
    LUCAS & 62596 & 400-2499 nm & \citep{lucas,lucas2015} \\
    ASSL & 19323 & 400-2499 nm & \citep{AUS} \\
    BRC & 1100 & 450-2499 nm & \citep{BRC} \\
    SER & 108 & 400-2499 nm & \citep{SER} \\
    OSSL & 23847 & 400-2499 nm & \citep{ossl} \\
    UKC & 105 & 400-2499 nm & \citep{UKC} \\
    BSSL & 16094 & 400-2499 nm & \citep{BSSL,BSSLa} \\
    BASE & 695 & 400-2499 nm & \citep{BASE} \\
    KEW & 553 & 850-2499 nm & \citep{KEW} \\
    OSSLJ & 48503 & 950-2499 nm & \citep{sanderman2023nir} \\
    AFSIS & 1907 & 850-2499 nm & \citep{AfSIS} \\
    BEL & 83 & 400-2449 nm & \citep{BES} \\
    BRT & 102 & 450-2149 nm & \citep{BRT} \\
    FRT & 1415 & 1000-2499 nm & \citep{FRdata1, FRdata2} \\
    ITR & 300 & 400-2499 nm & \citep{ITR} \\
    DAE & 26 & 400-2499 nm & \citep{DAE} \\
    GER1 & 362 & 1350-2499 nm & \citep{GER} \\
    GER2 & 362 & 400-2199 nm & \citep{GER} \\
  \hline
\end{tabular}
\label{tab:data4train}
\end{table}

\hspace*{-7cm}
\begin{table}
\centering
\caption{Datasets for model testing}
\begin{tabular}{>{\raggedright\arraybackslash}p{2cm} >{\raggedright\arraybackslash}p{1.5cm}>{\raggedright\arraybackslash}p{3.3cm}>{\raggedright\arraybackslash}p{1.6cm}>{\raggedright\arraybackslash}p{2cm}>{\raggedright\arraybackslash}p{2cm}}
  \hline
  Dataset & Number of samples & Valid soil property & Average number of soil property & Manufacturer of spectrometer & Reference\\
  \hline
    Barthès2023 & 404 & Clay, Silt, Sand, Organic carbon, and Total nitrogen contents, Coarse fragment, and Bulk density & 4.94 & FOSS & \citep{Barthès2023} \\ 
    Hu2020 & 47 & Clay, Silt, Sand, Mg, Al, P, Ca, Mn, Fe, Zn, and Ni contents & 10.47 & Spectral Evolution & \citep{hu2020} \\ 
    MARMIT2020 & 201 & Clay, Silt, Sand, Soil organic matter, CaCO$_3$, Organic carbon, and Total iron contents, Bulk density, pH value, and Cation-exchange capacity & 4.96 & ASD & \citep{marmit,marmit2} \\ 

  \hline
\end{tabular}
\label{tab:data4test}
\end{table}

\begin{table}
\centering
\caption{Soil property examples}
\begin{tabular}{>{\raggedright\arraybackslash}p{6cm} >{\raggedright\arraybackslash}p{3cm}>{\raggedright\arraybackslash}p{2.5cm}}
  \hline
  Property & Value example & Unit \\
  \hline
    Coarse fragments & 1.0 & \% \\
    Clay & 1.0 & \% \\
    Clay & 1.0 & $g/dm^3$ \\
    Silt & 1.0 & \% \\
    Sand & 1.0 & \% \\
    Cation exchange capacity & 1.0 & cmol(+)/kg \\
    Cation exchange capacity & 1.0 & cmol(+)/$dm^3$ \\
    pH measured from CaCl$_2$ solution & 1.0 & - \\
    pH measured from water solution & 1.0 & -   \\
    Total carbon content & 1.0 & \% \\
    Organic carbon content & 1.0 & \% \\
    The primary land cover & Sunflower & - \\
    The primary land use & Forestry & - \\
    Exchangeable phosphorus content & 1.0 & mg/kg \\
    Extractable phosphorus content & 1.0 & mg/kg \\
    Total phosphorus content & 1.0 & mg/kg \\
    Total potassium content & 1.0 & mg/kg \\
    Total nitrogen content & 1.0 & g/kg \\
    CaCO$_3$ content & 1.0 & g/kg \\
    Percentage of stones in soil & $<$10 & \% \\
    Country & United States & - \\
    Province/State & California & - \\
  \hline
\end{tabular}
\label{tab:prop}
\end{table}

\subsection{Spectra padding model}\label{S:model:pad}

The goal of the spectra padding sub-model is to standardize all spectra to a uniform wavelength interval spanning 400 to 2499 nm at 1 nm intervals, with consistent units. All raw spectral units were converted to wavelength, and spectral absorbance ($\mathbf{A}$) values were converted to reflectance ($\mathbf{R}$) according to the equation: $\mathbf{R} = 1/10^\mathbf{A}$. If reflectance values were reported at an interval greater than 1 nm, linear interpolation was used to determine values at a 1 nm interval. The spectra were then organized into a matrix of size $N\times 2100$, where $N$ is the total number of spectra in the dataset. 

As described before, the soil spectral libraries obtained by different spectroscopic instruments may have different measurement wavelength ranges. If a spectrum had wavelengths ranging from 400 to 2499 nm, it will occupy the whole matrix row and no padding is needed. If its range is less than the maximum range, the missing values in the matrix row will initially be set to 0. Wavelengths exceeding the maximum range of 400 to 2499 nm are removed. An input spectrum with a shorter range will be expanded to span the entire 400 to 2499 nm using the spectra padding model.

\begin{figure}
\includegraphics[width=\textwidth]{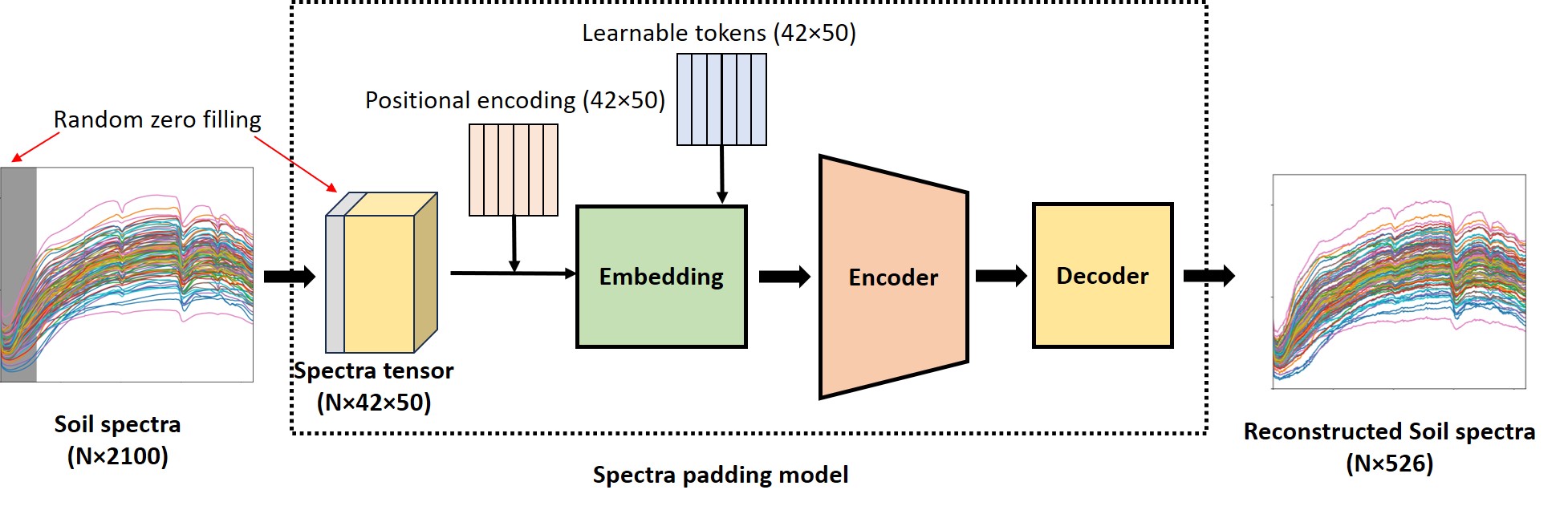}
\caption{Schematic representation of the training and architecture of the spectra padding model. A portion of the input spectra is randomly set to 0 (indicated by the grey shadow), and the down-sampled, full-range input spectra are reconstructed. The embedding block consists of transformer layers, while the encoder and decoder blocks are composed of 1D CNN layers.}
\label{fig:padmodel}
\end{figure}

The first embedding part of the spectra padding model is developed based on the encoder component of the soil spectra model of \citep{lei2022achieving}. The vector of a spectrum was reshaped into a matrix with a shape of $42\times 50$, where 42 is the number of wavebands, and 50 is the number of absorbance values in one waveband. In cases where the first or last waveband in the matrix is not fully filled, it is entirely set to zero. Consequently, the input spectral matrix was converted into a three-dimensional (3D) tensor (spectra tensor in Fig. \ref{fig:padmodel}).

Three transformer layers are then used to process this input data, designed so that the padding does not influence the calculations. In other words, the output of the model should be the same for tensors with any number of zero padding wavebands as long as the non-zero wavebands are the same. Therefore, a masked transformer was used to eliminate the effect of zero padding. The first two transformer layers use a self-attention mechanism, and if the input band is zero-padded, the same index of the output will also be zero-padded. As the attention mechanism cannot recognize the order of input wavebands, positional encoding that adds position information of the tokens in the sequence is suggested \citep{attention, lei2022achieving}. The positional encoding has the same dimension as the input spectra matrix so that the two can be summed. It should be noted that positional encoding is added only to non-zero bands. To make sure the outputs have the same shape without zero padding bands, the third transformer layer is based on a cross-attention mechanism. Specifically, 42 learnable tokens are used as query input of the third layer, and the output of the second layer is used as key and value input. Then, the output of spectral embedding enters an encoder and then a decoder. The encoder and decoder consist of one-dimensional (1D) convolution neural networks (CNN).

The input spectra undergo zero padding on randomly selected wavebands (Fig. \ref{fig:padmodel}). The target is the down-sampled input spectra without zero padding, still ranging from 400 to 2499 nm, but with a resolution of 4 nm, resulting in an output matrix of $N\times 526$. This down-sampling aims to accelerate model training and prevent overfitting.

The trained model is then capable of padding the missing wavelengths of input soil reflectance spectra. The output down-sampled spectra are up-sampled to align with the size of the input spectra ($N\times 2100$) using linear interpolation. The non-padded parts of the up-sampled spectra are replaced with the original spectra. Lastly, a second-order Savitzky-Golay smoothing \citep{sgsmooth}, with a window size of 100 nm, is applied to the padded parts of the spectra, resulting in the final full-wavelength spectra.

\subsection{Property embedding}\label{S:pe}

In the present study, we input text strings into the SOGM that include full soil property descriptions rather than just numerical values and labels. The properties from all the collected datasets were organised in the following format: ``property: value unit", with some examples shown in Table \ref{tab:prop}. This table reveals connections among various soil properties. Subsequently, a learnable word dictionary containing parameterized vectors was built. The total number of vectors corresponds to the unique words present across all properties, meaning each vector represents a specific word. Consequently, a phrase like ``Exchangeable phosphorus content: 1.0 mg/kg" can be converted into a matrix. Information about the spectrometer manufacturer is also input into the SOGM in the same format as other soil properties, for example, ``Spectrometer manufacturer: ASD". Numerical values in sentences are represented by the actual number along with a numerical value tag vector, and this tag is also included in the word dictionary.

Positional encodings are added to the sentence matrix to enable the model to identify the sequence of words. This matrix is then aggregated into a sentence embedding vector using transformer layers. These sentence embeddings are further combined to create a property embedding, employing additional transformer layers. However, positional encodings are not used for sentence embeddings, as soil properties do not possess an inherent order. The transformer layers produce consistent results regardless of the sentence embeddings' order, provided the input sentence embeddings remain the same. All transformer layers in the property embedding model are equipped with masks, similar to those in Sect. \ref{S:model:pad}, to accommodate varying sentence and word counts.

A predefined list, featuring text descriptions for 15 major soil properties, is provided alongside the codes. Users can consult the code for valid input values.

\subsection{Denoising diffusion}\label{S:ddpm}

The denoising diffusion probabilistic model (DDPM) proposed by \citep{ddpm} was developed for high-quality image generation, which is a class of latent variable models inspired from nonequilibrium thermodynamics. The diffusion process begins by gradually adding noise into a clear image over a series of time steps, effectively transforming the image into a noisy version. The specific time step indicates the stage within the diffusion process. Then, a U-Net model \citep{unet} is trained to predict the noise at each time step and subtract it from the noisy image. Through this iterative denoising, the model effectively 'reverses' the diffusion process. The key is that through learning to reverse this noise addition, the DDPM gradually reconstructs the original or a new image from the noisy state. 

\begin{figure}
\includegraphics[width=\textwidth]{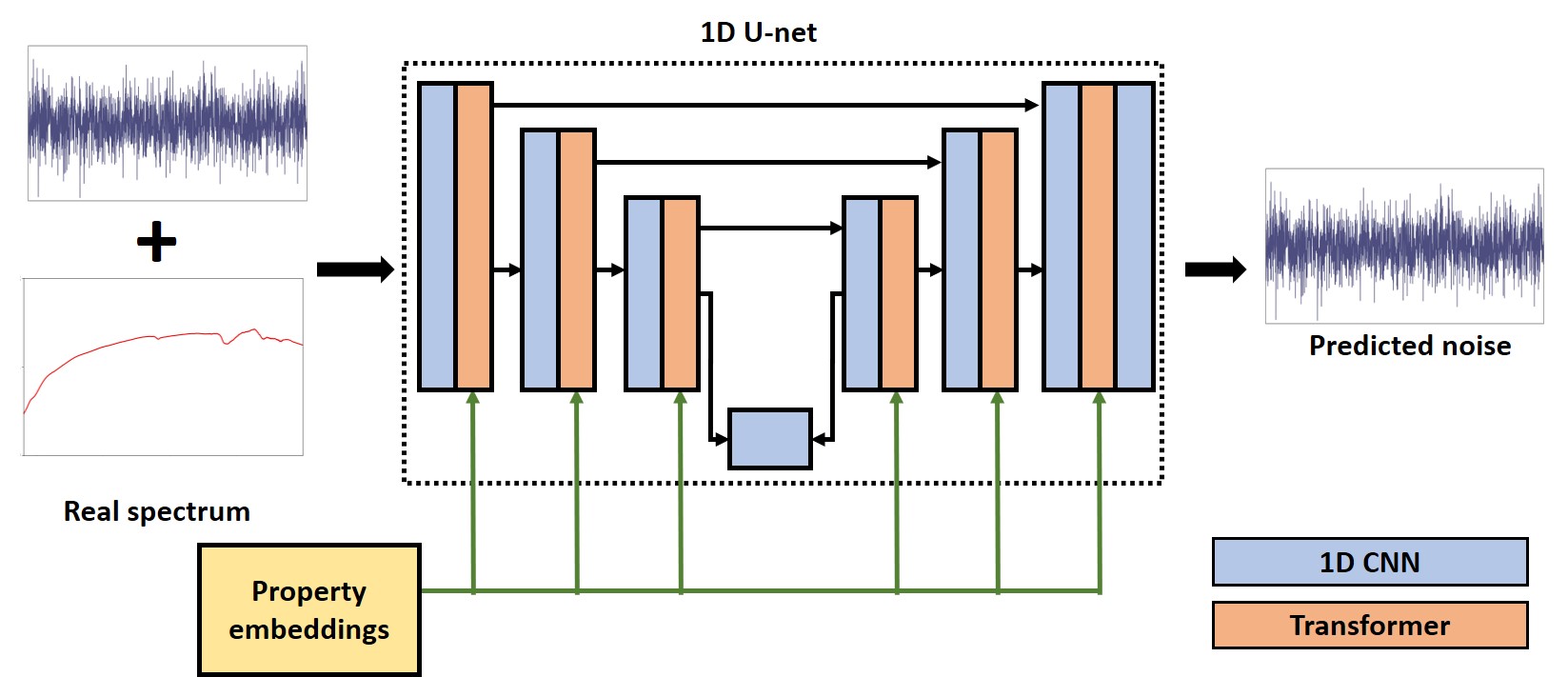}
\caption{Schematic representation of the denoising diffusion model. The inputs consist of the sum of random noise and soil spectra, along with corresponding property s. The output is the input random noise during model training. The blue blocks represent 1D CNN layers, and the orange blocks represent transformer layers.}
\label{fig:ddpmodel}
\end{figure}

The text embeddings obtained by the property embedding model were integrated into the noise prediction by adding a transformer layer into the U-Net (Fig. \ref{fig:ddpmodel}). A 1D U-Net, which consists of 1D CNN layers, was employed due to the 1D nature of soil reflectance spectra data. For model training, random noise vectors were sampled and added to the spectra in different proportions as determined by the time step. The property text strings are input into the property embedding model to obtain the embeddings. Subsequently, the noisy spectra and corresponding property embeddings are fed into the U-Net model to predict the added noise. The property embedding model and the diffusion model are trained together to minimize the loss function $\mathcal{L}$, which is given by:

\begin{equation}
\mathcal{L} = \mathbb{E}_{t, \mathbf{R}_0, \epsilon}\left[ \min(\mathrm{SNR}(t),\lambda)\|\epsilon - \epsilon_{\theta}(\mathbf{R}_t, t)\|^2 \right],
\end{equation}

\noindent where $\mathbb{E}$ is the expectation over the random variables, $t$ denotes the time step in the diffusion process,	$\mathbf{R}_0$ is the original soil spectrum, $\epsilon $ is the true noise added at time $t$, $\epsilon_{\theta}(\mathbf{x}_t, t)$ refers to the noise predicted by the model, given the noisy spectrum $\mathbf{R}_t$ at time $t$, $\|\cdot\|^2 $ denotes the squared Euclidean norm, used here to calculate the mean squared error, the factor $\lambda$ is set to 5, and $\mathrm{SNR}(t)$ represents the signal-to-noise ratio at time $t$. It can be calculated by: $\mathrm{SNR}(t) =\frac{\bar{\alpha}_t}{1 - \bar{\alpha}_t}$, where $\bar{\alpha}_t$ is the product of noise scales $\alpha$ from the start of the diffusion process up to time $t$. The use of $\mathrm{SNR}(t)$ is to avoid the model focusing too much on small noise levels \citep{hang2023efficient}.

\begin{figure}
\includegraphics[width=\textwidth]{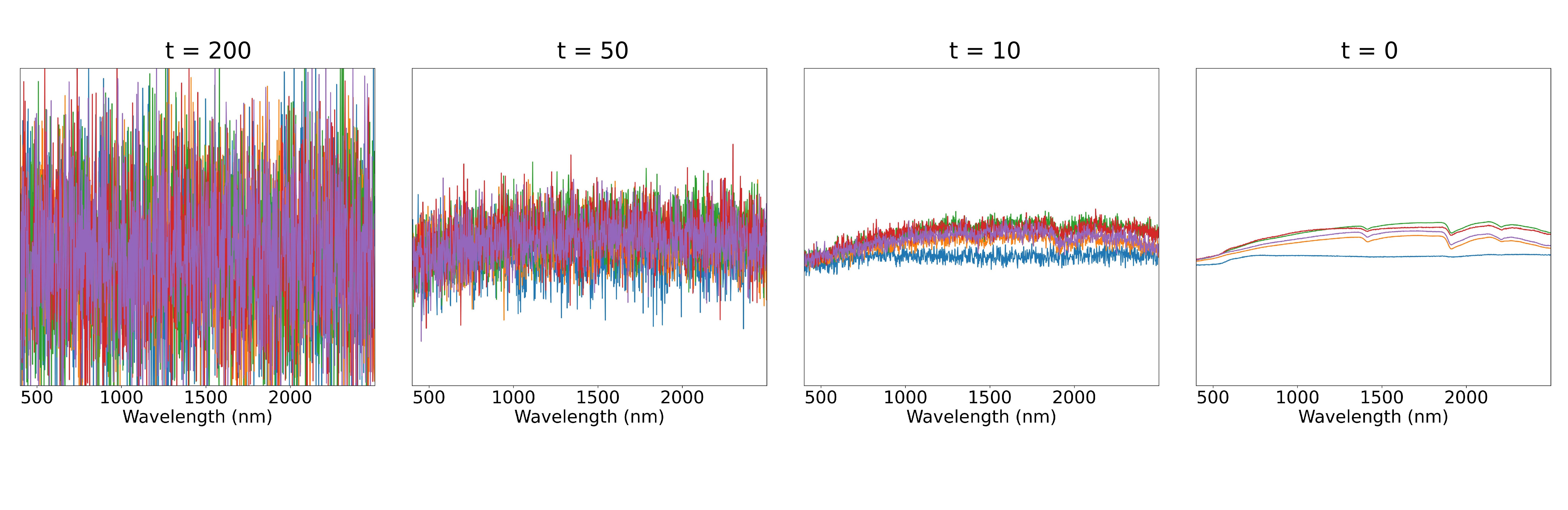}
\caption{Example illustration of the spectral denoising process. During the reverse diffusion process, the soil spectra are progressively recovered as the time step decreases. Four example spectra are shown at time steps ($t$): 200, 50, 10, and 0.}
\label{fig:denoise}
\end{figure}

Once trained, the model generates new spectra by reversing the diffusion process controlled by the target soil property text description. It starts with a sample of random noise and then iteratively applies the learned reverse transformations to reduce the noise. Each step in the reverse process involves applying the model to predict and subtract out the noise from the current spectrum, effectively denoising it (Fig. \ref{fig:denoise}). The total number of time steps was set to 300, which was empirically found to consistently ensure adequate denoising efficiency and quality.

\subsection{Wet soil spectra model}\label{S:water}

The gravimetric soil moisture content ($SMC_g = \frac{100 \times (m_w - m_d)}{m_d}
$, where $m_w$ and $m_d$ are wet and dry soil weights for a given sample volume, respectively) is an external factor affecting the reflectance spectrum, but typically changes much more rapidly than other properties. Large soil spectral libraries are based on dry soil, thus the present SOGM can only generate dry soil reflectance spectra as all training data only contain dry soil spectra. For modelling the soil reflectance as a function of the soil moisture content, a separate model was introduced.

The reflectance spectra of wet soil are typically a series corresponding to discrete soil moisture contents derived from the reflectance spectra of dry soil measurements \citep{marmit,marmit2,tian2021soil}. Thus a regression model based on a 1D U-Net model \citep{unet} was used to determine the wet soil spectrum when the dry soil spectrum and soil moisture content values are provided. The output data of the model is the difference between dry and wet soil spectra. From this, the wet soil spectra can be obtained by subtracting this output from the dry spectra. Therefore, unlike the SOGM which can use flexible combinations of soil properties as input, the wet spectra model requires specific input data (i.e., $SMC_g$ and reference dry spectrum), making it a deterministic model without the variability inherent in the SOGM.

To align with the requirements of the model, the data from MARMIT2020 \citep{marmit,marmit2} and Tian2021 \citep{tian2021soil} repositories were converted into 1670 samples, each of which contain a dry soil spectrum, a SMC value, and the corresponding wet soil spectrum. The dataset was randomly divided into 1300 training spectra, and 370 testing spectra.

\subsection{Soil-plant image generation based on 3D ray-tracing model}\label{S:RT}

Synthetic images of model scenes containing plants with soil background were generated using the Helios 3D plant modeling software (v1.3.0) \citep{HELIOS}. The Helios software enables generation and manipulation of fully-resolved 3D geometric models of plants, the ground, or other objects. For example, plant canopy geometries of bean crops used in the present study were created using the Helios ``Canopy Generator" plug-in. The procedural models have user-defined geometric parameters such as canopy height, leaf size, leaf area index (LAI), and leaf angle distribution, which allows for easy customization of the specifics of the canopy, and can be used as annotations for output images at both large and small scales. Users can also add external model geometries to the scene from standard polygon file formats. Geometry added to the scene can be referenced based on their unique identifiers in order to assign their spectral radiative properties.

\begin{figure}
\includegraphics[width=\textwidth]{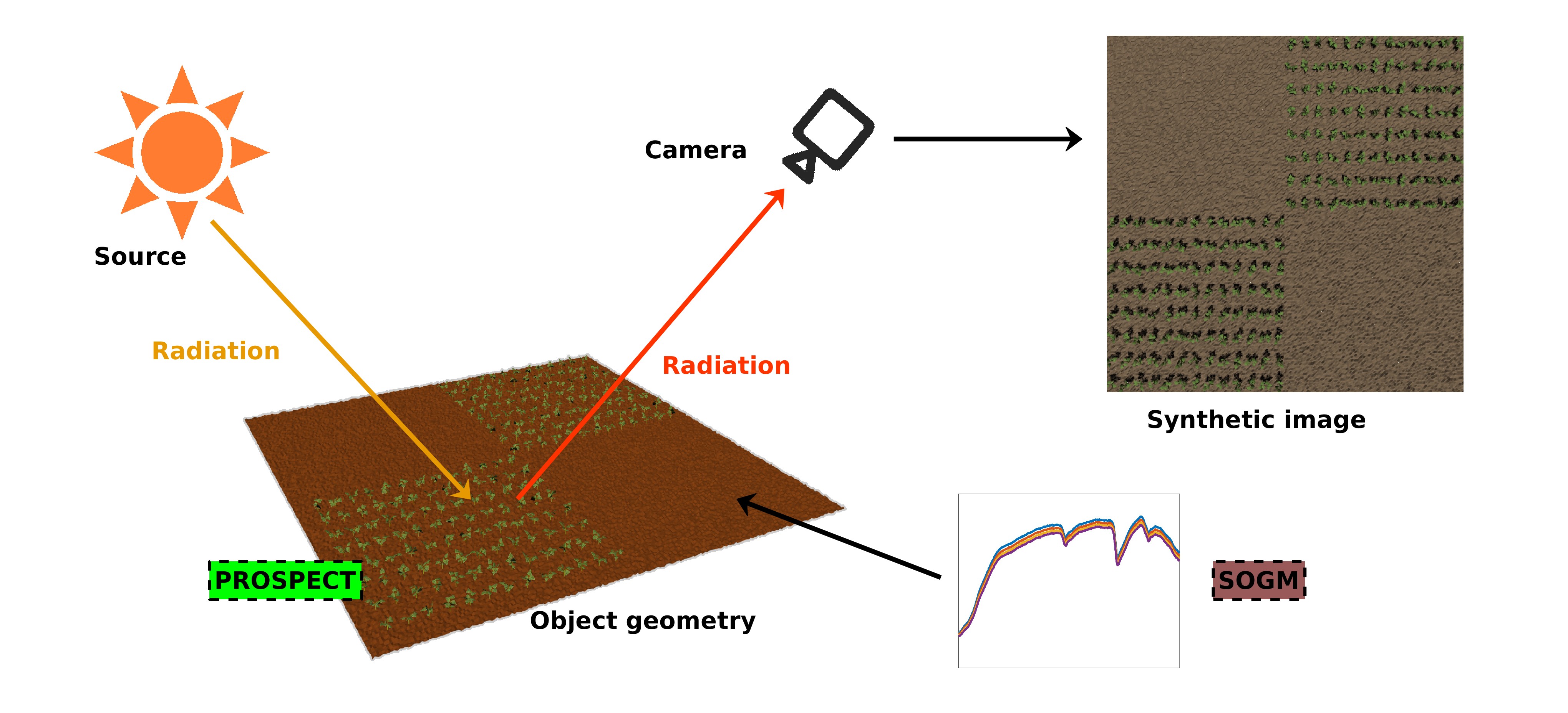}
\caption{Schematic representation of the 3D radiation model for image generation.  A ray-tracing-based camera model is used to simulate radiation that is emitted from a radiation source (e.g., sun, LED light) and reaches the camera after being scattered by objects in the scene. The PROSPECT-based leaf optical model and SOGM can generate the leaf and soil optical properties, respectively. Finally, the simulated camera generates resulting images that can be arbitrarily auto-annotated.}
\label{fig:rt}
\end{figure}

In order to generate synthetic images, the distribution of absorbed, reflected, transmitted, and emitted radiation for all primitive elements in the scene is computed for a single scattering iteration based on the 3D ray-tracing method proposed by \citep{RAYTRACYING}. A ray-tracing-based camera model is then used to sample the reflected and transmitted energy for every camera pixel across all wave bands (Fig. \ref{fig:rt}). Scattering iterations continue for multiple scattering instances, and the camera continues accumulating scattered radiation until the amount of remaining scattered radiation becomes arbitrarily small. The camera also uses ray-tracing to determine primitive elements contained within each pixel, which is then used for image labeling. The radiation transport among objects and the radiation received by the simulated camera sensor has been verified by using the RAMI On-line Model Checker \citep{ROMC} (ROMC, \url{https://romc.jrc.ec.europa.eu/_www/}).

Before running the ray tracing model, the optical properties, including the total hemispherical reflectivity $\rho$ and transmissivity $\tau$ across each radiative band, must be assigned to the corresponding geometric elements within the scene. These properties also incorporate the camera spectral response for the particular band. They can be calculated as:

\begin{align}\label{eq:integ_ro}
\rho_o = \frac{\int_{\lambda_{min}}^{\lambda_{max}}\,\rho_\lambda\,C_\lambda\,S_\lambda\,d\lambda}{\int_{\lambda_{min}}^{\lambda_{max}}\,S_\lambda\,d\lambda}, 
\end{align}
\begin{align}\label{eq:integ_to}
\tau_o = \frac{\int_{\lambda_{min}}^{\lambda_{max}}\,\tau_\lambda\,C_\lambda\,S_\lambda\,d\lambda}{\int_{\lambda_{min}}^{\lambda_{max}}\,S_\lambda\,d\lambda}, 
\end{align}

\noindent where $\lambda$ refers to the wavelength, and $\lambda_{min}$ and $\lambda_{max}$ represent the lower and upper bounds, respectively, of the selected waveband, $C_\lambda$ is the spectral sensitivity of the camera sensor for wavelength $\lambda$ (the whole camera spectral response is calibrated according to \ref{AppB}), and $\rho_\lambda$, $\tau_\lambda$, and $S_\lambda$ are the spectral reflectivity, spectral transmissivity, and spectral source flux at wavelength $\lambda$, respectively. The $\rho_\lambda$ of a soil primitive is one value in the soil reflctance spectrum $\mathbf{R}_0$ at wavelength $\lambda$ generated by the present SOGM (Fig. \ref{fig:rt}). The $\rho_\lambda$ and $\tau_\lambda$ of leaves can be generated by the PROSPECT-D \citep{PROSPECTD} or PROSPECT-PRO model \citep{PROSPECTPRO}, which has been implemented into Helios. Alternatively, $\rho_\lambda$, $\tau_\lambda$, and $S_\lambda$ can be manually measured by using spectroscopic devices that cover the VIS-NIR range.

The model allows automatic annotation of generated images based on an arbitrary geometric grouping or any variable simulated in Helios. The annotation process starts by determining the unique identifiers of geometric element(s) contained in every pixel of the simulated image. Once these identifiers are known, any information about these primitives available in Helios (e.g., type, angle, area, computed fluxes) can be queried and used to generate ``labelled" images.

\subsection{Model performance evaluation}\label{S:verf}

To verify the performance of the SOGM, the soil properties from testing datasets were input into the SOGM, and the generated spectra were compared against the corresponding real spectra from the testing datasets. As the generative model has uncertainty, the model was run 10 times with different random seeds, from which mean spectra were calculated to get relatively stable results for model evaluation. The generated mean spectra were then used for final evaluation. The testing spectra were padded using the present spectra padding model before comparison, and the performance of the spectra padding model was also evaluated. The wet soil spectra model was also evaluated by comparing the real and predicted wet soil spectra based on input SMCs and corresponding dry soil spectra. The root mean square error ($\mathrm{RMSE}$) averaged across all spectra in the dataset was used to evaluate the absolute errors, as the values of reflectivity is an important physical parameter in radiation transfer modelling. The mean square of Pearson's correlation coefficient ($r^2$) was used to evaluate the correlation between predicted and real spectra. The mean $\mathrm{RMSE}$ and $r^2$ can be calculated as:

\begin{equation}
       \mathrm{RMSE} = \frac{1}{n}\sum_{i=1}^{n}\mathrm{RMSE}_i,
\end{equation}

\begin{equation}
       r^2 = \frac{1}{n}\sum_{i=1}^{n}r_i^2,
\end{equation}

\noindent where $\mathrm{RMSE}_i$ and $r_i^2$ are the root mean square error and square of correlation coefficient of the $i$-th pair of generated and real spectra, respectively.

The soil colors in generated soil images were also evaluated based on a photo from \citep{sadeghi2018particle}. The mean color RGB values of all pixels for one type of soil from real and synthetic soil images were used for comparison. As the light and camera details are not provided in the original literature (which are required inputs for the camera model), only $r^2$ was used for evaluation as it gives a relative rather than absolute measure of agreement.

\section{Results}

\subsection{Spectra padding}

The wavelength range of the Barthès2023 and Hu2020 datasets spans from 1100 to 2499 nm and 400 to 2499 nm, respectively. The wavelength range of MARMIT2020 \citep{marmit,marmit2} extends from 400 nm to approximately 2349 - 2449 nm, as it contains spectral data from multiple sources. The missing wavelengths between 400 and 2499 nm were reconstructed using the present spectra padding model. 

\begin{figure}
\includegraphics[width=\textwidth]{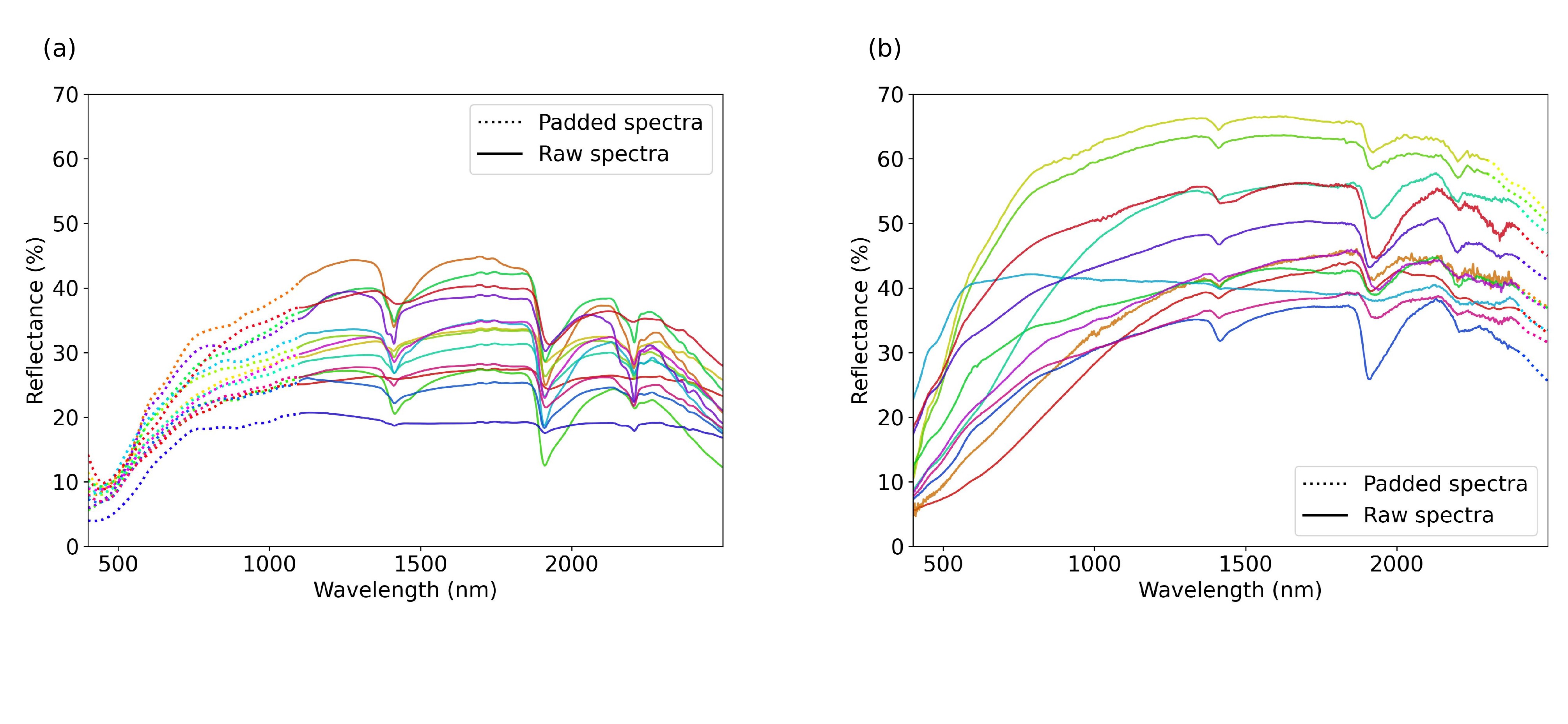}
\caption{Full-range spectra obtained by applying the spectra padding model to (a) Barthès2023 and (b) MARMIT2020 datasets. The solid curves represent the original spectra, and the dotted curves are the reconstructed portion of the spectra.}
\label{fig:pad}
\end{figure}

Figure \ref{fig:pad} shows some padded spectra from Barthès2023 and MARMIT2020 data sets, and the padding overall looks reasonable based on visual inspection. Nevertheless, the accuracy of these padding results cannot be evaluated, as the missing parts were not provided by the original datasets. Thus, to verify the performance of the spectra padding model, spectral bands 400-799 nm, 400-1099 nm, and 2100-2499 nm were set to 0, and the model was used to predict these zero-set bands based on the remaining values. Table \ref{tab:padding_result} displays the error metrics for all three selected wavebands, which indicated good performance for both the Hu2020 and MARMIT2020 datasets ($\mathrm{RMSE}<3$ and $r^2>0.8$). It should be noted that not all spectral bands within 2100-2499 nm in MARMIT2020 dataset are evaluated, as some spectra do not fully cover this range. These results of 400-1099 nm and 2100-2499 nm indicate the reconstructed missing parts of Barthès2023 and MARMIT2020 (Fig. \ref{fig:pad}) are reliable. The spectral bands 400-799 nm and 400-1099 nm are missing in the original Barthès2023 dataset, thus these results are unavailable in Table \ref{tab:padding_result}. The $\mathrm{RMSE}$ for the 2100-2499 nm range in the Barthès2023 dataset is higher than that of the other two datasets.  This is reasonable, as it only utilizes the spectral band from 1100-2099 nm (1000 wavelengths) to reconstruct the missing part, which is fewer than the wavelengths (1700) used in the other two datasets.

\hspace*{-7cm}
\begin{table}
\centering
\caption{Evaluation of spectra padding model based on spectral datasets listed in Table \ref{tab:data4test}. Reflectance values were set to 0 across 3 target wavebands, and the padding model was used to reconstruct the zeroed values.}
\begin{tabular}{>{\raggedright\arraybackslash}p{2.8cm}>{\raggedright\arraybackslash}p{1.2cm}>{\raggedright\arraybackslash}p{1.2cm}>{\raggedright\arraybackslash}p{1.2cm}>{\raggedright\arraybackslash}p{1.2cm}>{\raggedright\arraybackslash}p{1.2cm}>{\raggedright\arraybackslash}p{1.2cm}}
  \hline
    Target waveband & Barthès2023 & & Hu2020 & & MARMIT2020 & \\
     & $\mathrm{RMSE}$ & $r^2$ & $\mathrm{RMSE}$ & $r^2$ & $\mathrm{RMSE}$ & $r^2$ \\
  \hline
    400-799 nm & - & - & 1.68 & 1.00 & 2.81 & 0.96 \\
    400-1099 nm & - & - & 1.95 & 1.00 & 3.07 & 0.96 \\
    2100-2499 nm & 3.54 & 0.66 & 2.18 & 0.92 & 2.06 & 0.81 \\
  \hline

\end{tabular}
\label{tab:padding_result}
\end{table}

\subsection{Spectra generation}
\begin{figure}
\includegraphics[width=\textwidth]{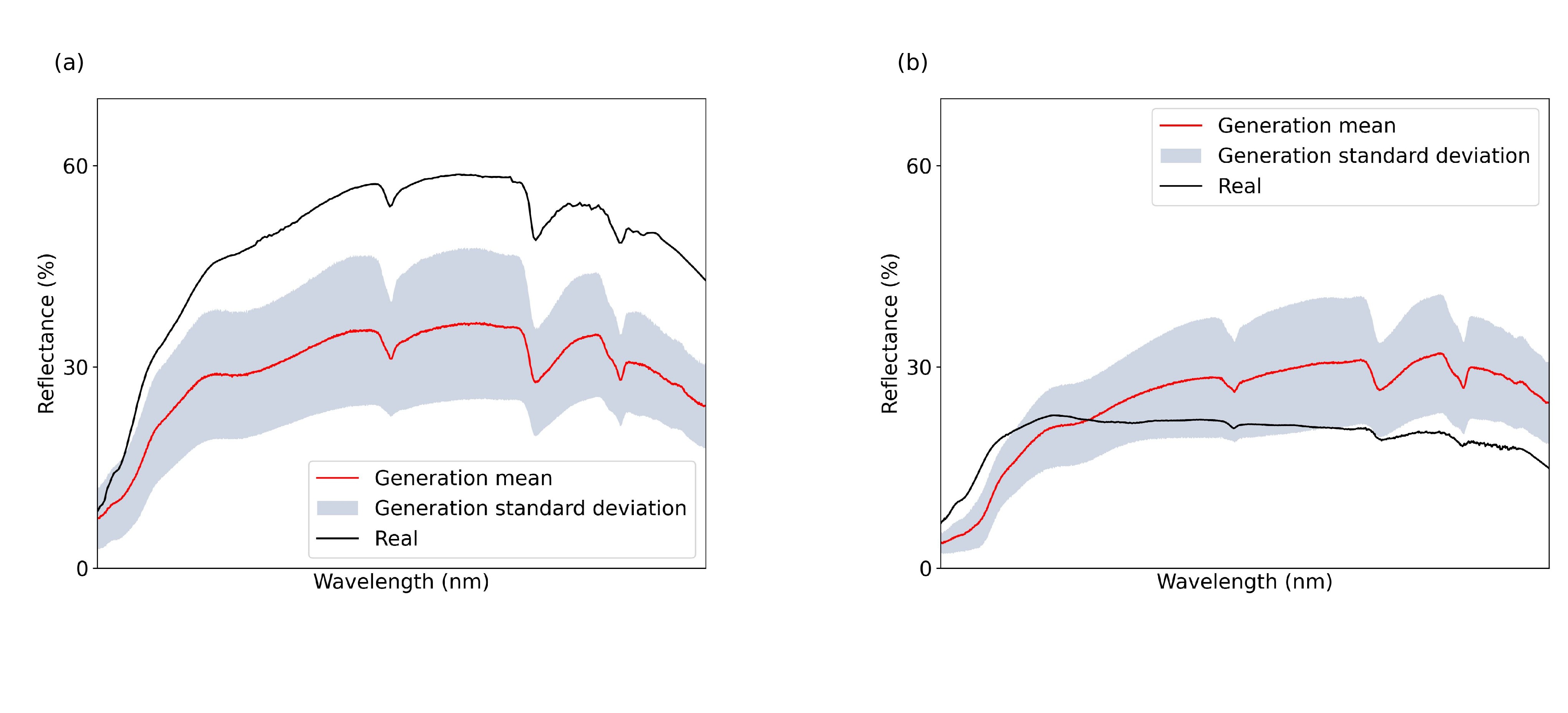}
\caption{Example real soil reflectance spectra and mean and standard deviation of generated soil reflectance spectra with similar soil particle fraction properties and spectrometer manufacturer (ASD). (a) Clay : 22.0\%; Silt : 26.0\%; Sand : 52.0\%. (b) Clay : 17.3\%; Silt : 26.1\%; Sand : 56.6\%.}
\label{fig:particlespec}
\end{figure}

When all available input properties are provided to the model, the $\mathrm{RMSE}$ for the Barthès2023 dataset was 5.52\% (Table \ref{tab:data4test_result}) with an $r^2$ of 0.86. Excluding the input information of spectrometer manufacturer leads to larger errors for all three datasets, which means the model effectively learned the spectral variation caused by different spectrometer manufacturers from the training datasets without ever seeing these test datasets. For datasets Barthès2023 and MARMIT2020, randomly dropping one or two input properties led to increased errors, which is intuitively expectated. Figure \ref{fig:particlespec} shows spectra from two example soil samples with similar clay, silt, and sand fractions, yet they exhibit very different optical properties. Consequently, if only these properties are input into the SOGM, the model can predict a spectrum that is ``reasonable", but more information is needed to accurately describe the soil spectra. In the case of Barthès2023, the $\mathrm{RMSE}$ is 5.92\% when only particle size information (clay, silt, and sand fraction) is input. Adding nitrogen and organic carbon improves the $\mathrm{RMSE}$ to 5.19\% and 5.69\%, respectively, as shown in Table \ref{tab:data4test_result}.

\hspace*{-7cm}
\begin{table}
\centering
\caption{Error in soil spectra generative model predictions based on variable combinations of input properties. The model was trained based on the spectral datasets listed in Table \ref{tab:data4train}, and then used to predict the spectral datasets Barthès2023, Hu2020 and MARMIT2020 (Table \ref{tab:data4test}) based on variable input parameters. Error between predicted and true spectral reflectance was quantified by $\mathrm{RMSE}$ (\%) and $r^2$. $n$ denotes the number of spectra considered.}
\begin{tabular}{>{\raggedright\arraybackslash}p{2.5cm}>{\raggedright\arraybackslash}p{0.8cm}>{\raggedright\arraybackslash}p{0.8cm}>{\raggedright\arraybackslash}p{0.8cm}>{\raggedright\arraybackslash}p{0.8cm}>{\raggedright\arraybackslash}p{0.8cm}>{\raggedright\arraybackslash}p{0.8cm}>{\raggedright\arraybackslash}p{0.8cm}>{\raggedright\arraybackslash}p{0.8cm}>{\raggedright\arraybackslash}p{0.8cm}}
  \hline
  Dataset & Barthès2023 & & & Hu2020 & & & MARMIT2020 & & \\
  \hline
    Property & $\mathrm{RMSE}$ & $r^2$ & $n$ & $\mathrm{RMSE}$ & $r^2$ & $n$ & $\mathrm{RMSE}$ & $r^2$ & $n$ \\
  \hline
    All & 5.52 & 0.86 & 404 & 12.91 & 0.90 & 47 & 13.54 & 0.92 & 201 \\
    - Manufacturer & 6.20 & 0.81 & 404 & 13.34 & 0.92 & 47 & 14.17 & 0.90 & 201 \\
    - 1 & 6.48 & 0.79 & 404 & 11.43 & 0.94 & 47 & 13.98 & 0.92 & 201 \\
    - 2 & 6.60 & 0.79 & 404 & 11.71 & 0.94 & 47 & 14.69 & 0.90 & 201 \\
    Particle & 5.92 & 0.90 & 229 & 9.98 & 0.97 & 46 & 14.92 & 0.90 & 185 \\
    OC & 6.26 & 0.85 & 404 & - & - & - & 11.56 & 0.92 & 99 \\
    Particle \& OC & 5.69 & 0.90 & 229 & - & - & - & 11.69 & 0.94 & 92 \\
    Fe & - & - & - & 12.08 & 0.94 & 47 & 12.01 & 0.88 & 104 \\
    Particle \& Fe & - & - & - & 9.76 & 0.97 & 46 & 10.63 & 0.92 & 101 \\
    Particle \& Fe \& Mg & - & - & - & 9.69 & 0.97 & 46 & - & - & - \\
    Particle \& Fe \& Ni & - & - & - & 12.39 & 0.94 & 40 & - & - & - \\
    Density & 6.96 & 0.77 & 404 & - & - & - & 10.35 & 0.92 & 16 \\
    Particle \& Density & 5.96 & 0.85 & 229 & - & - & - & 14.94 & 0.97 & 2 \\
    N & 6.11 & 0.81 & 404 & - & - & - & 12.27 & 0.94 & 3 \\
    Particle \& N & 5.19 & 0.90 & 229 & - & - & - & 12.66 & 0.90 & 2 \\
    SOM & - & - & - & - & - & - & 13.80 & 0.90 & 107 \\
    Particle \& SOM & - & - & - & - & - & - & 12.45 & 0.92 & 104 \\
    OC \& SOM \& CaCO$_3$ & - & - & - & - & - & - & 11.13 & 0.92 & 95 \\
    Particle \& OC \& SOM \& CaCO$_3$ & - & - & - & - & - & - & 11.14 & 0.94 & 92 \\

  \hline
\end{tabular}
\smallskip
\begin{flushleft}
\small Soil property input abbreviations: All: all available soil properties; -Manufacturer: all available properties except spectrometer manufacturer; -1 and -2: all available properties except one or two soil properties are randomly omitted (if omission leaves only one remaining property, no properties will be dropped); Particle: clay, sand, and silt contents; OC: organic carbon content; Fe: total iron content; Mg: total magnesium content; Ni: total nickel content; Density: bulk density; N: total nitrogen content; SOM: soil organic matter content.
\end{flushleft}
\label{tab:data4test_result}
\end{table}

Interestingly, for the dataset Hu2020, randomly reducing the number of input properties actually led to better results. This result is reasonable when the input properties of this dataset are considered, since this dataset contains many elements with low concentration ($<$0.1\%) such as nickel and zinc. These minority elements have relatively higher measurement error, and also appeared a few times in the training datasets. For these reasons, including such properties as inputs may have a negative effect on model performance. If only the particle fraction property is input into the model, the resulting $\mathrm{RMSE}$ of 9.98\% and $r^2$ of 0.97 is much better than when all input properties are included (Table \ref{tab:data4test_result}). Adding iron and magnesium inputs can further improve model performance, as these two elements have relatively high concentration around 1$\sim$5\%. However, adding nickel (content $<$0.005\%) tends to increase overall model error.     

\begin{figure}
\includegraphics[width=\textwidth]{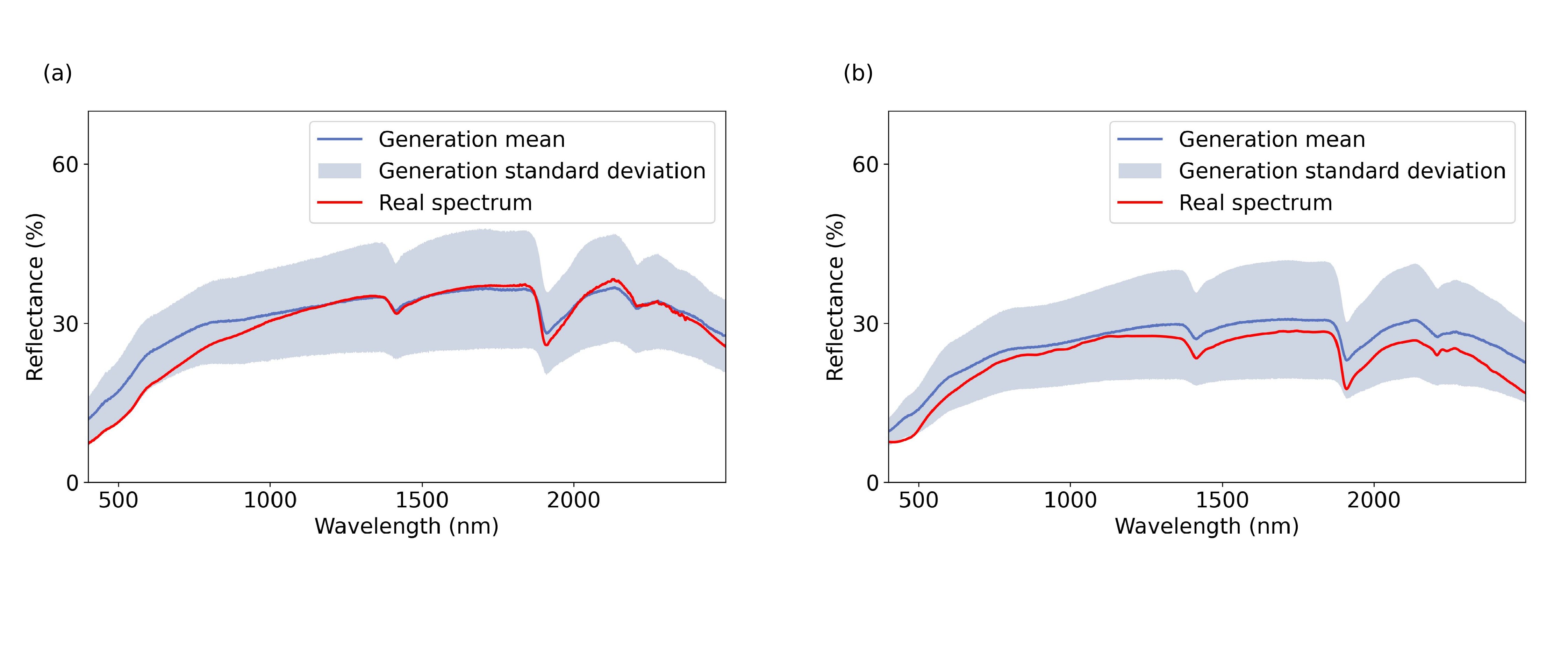}
\caption{Variability in predicted soil reflectance spectra based on incomplete inputs. Real soil spectra are compared against the distribution of 10 generated spectra based on soil properties of (a) Spectrometer manufacturer: ASD, Clay: 45.7 \%, Silt: 34.8 \%, Sand: 19.4 \%, Soil organic matter: 14.6 g/kg, CaCO3 content: 36 g/kg, Total iron content: 40700.0 mg/kg, Organic carbon content: 8.5 g/kg; (b) Spectrometer manufacturer: Spectral Evolution, Clay: 62.9 \%, Sand: 20.0 \%, Silt: 17.1 \%, Bulk density: 1.036 g/cm$^3$, Organic carbon content: 18.9 g/kg, Total nitrogen content: 1.57 g/kg. }
\label{fig:uncertainty}
\end{figure}

There is uncertainty and random variability in the SOGM spectra generation, which is a trade-off to the model's ability to predict reasonable spectra with incomplete inputs. Figure \ref{fig:uncertainty} presents real soil spectra alongside the distribution of 10 generated spectra based on the same property inputs. The generated mean spectra fall within an more acceptable range than generated mean spectra shown in Fig. \ref{fig:particlespec}, as more soil properties are provided. In general, averaging is recommended to reconstruct a stable spectrum. For other purposes, such as image generation or soil property estimation, averaging may not be necessary, as the generation uncertainty can contribute to greater data variation. Furthermore, by using the same random seeds, the model can also generate a series of soil spectra based on a gradient in soil properties.

\subsection{Wet soil spectra}

The soil moisture model achieved an $\mathrm{RMSE}$ of 3.19\% and $r^2$ of 0.90 on the 370 wet soil spectra testing set. While these errors are relatively low, the result can potentially be improved further by increasing the model parameters and number of training iterations. However, the number of spectra available for training the wet spectra model was much smaller than that used for the (dry) SOGM. Therefore, training was stopped at 5000 iterations to prevent overfitting. It was observed that additional training iterations could achieve lower errors on the testing set, but when the model trained with $>$5000 iterations was applied to unseen generated spectra, spectra with smaller moisture content could have lower overall reflectance than those with larger moisture content, which is not in line with reality.

\begin{figure}
\includegraphics[width=\textwidth]{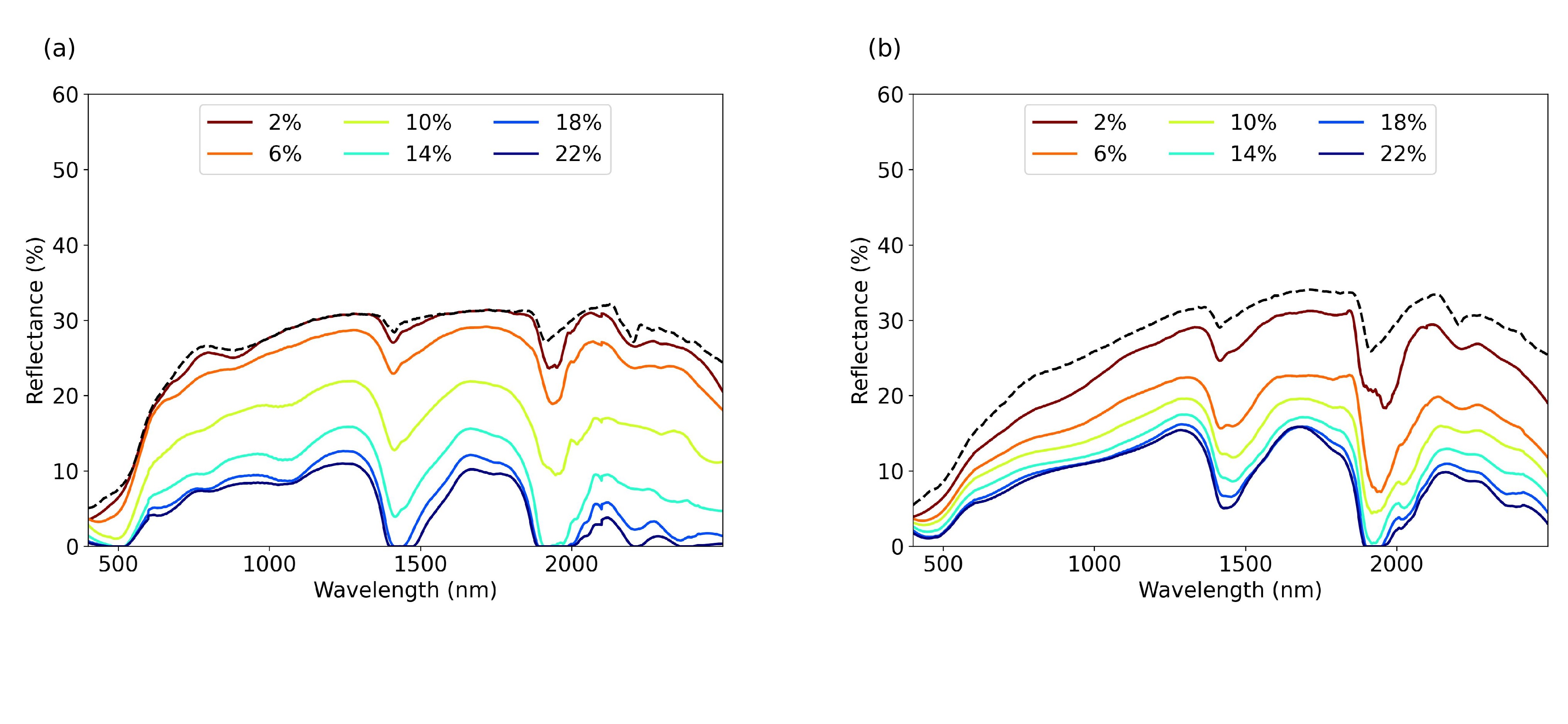}
\caption{Wet soil reflectance spectra predicted by the wet soil spectra model based on dry soil spectra (black dotted lines) randomly generated by the SOGM. The values in the figure legends are the input soil moisture content (\%).}
\label{fig:wetspec}
\end{figure}

The SOGM was then coupled with the wet soil model by using the dry spectra generated by the SOGM as input to the wet soil model (Fig. \ref{fig:wetspec}). Figure \ref{fig:wetspec} shows wet soil reflectance spectra predicted by the wet spectra model based on several input SMCs and two dry soil spectra randomly generated by the SOGM. Eventually, the integration of SOGM with the wet spectral model enables SOGM to generate wet soil spectra.

\subsection{Synthetic soil images}

\begin{figure}
\includegraphics[width=\textwidth]{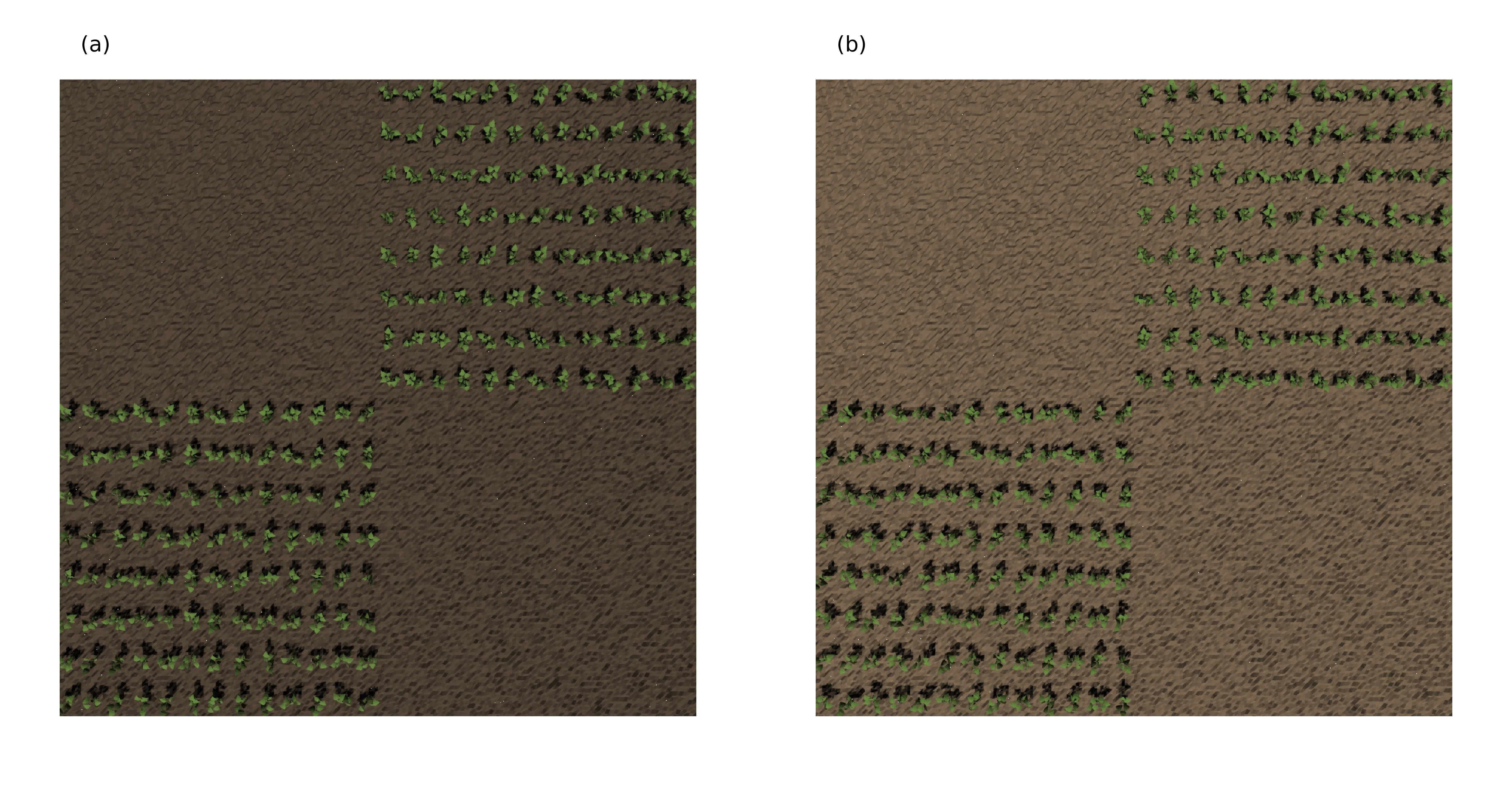}
\caption{Synthetic soil-plant images based on soil properties of (a) \textit{Terra Preta} soil: Clay: 8 \%, Silt: 7 \%, Sand: 85 \%, Soil organic matter: 120 g/kg, Total iron content: 80000 mg/kg, Organic carbon content: 38 g/kg,  Cation exchange capacity: 123 cmol(+)/kg, pH measured from water: 4.7; (b) soil near Davis, CA, USA: Clay: 33 \%, Silt: 47 \%, Sand: 20 \%, Soil organic matter content: 30 g/k, Organic carbon content: 8 g/kg, Electrical conductivity 55 mS/m, pH measured from water: 6.8, Cation exchange capacity: 30 cmol(+)/kg). These scenes were illuminated by simulated sun light using the ASTM standard clear sky solar spectrum distribution with solar zenith angle of 20\degree. A simulated Nikon B500 camera (Nikon, Minato City, Tokyo, Japan) was used for imaging, and the camera response spectra used is shown in Fig. \ref{fig:3spectra}b.}
\label{fig:soilimages}
\end{figure}

Figure \ref{fig:soilimages} shows synthetic soil-plant images generated using the Helios 3D plant modelling software \citep{HELIOS}. The soil reflectance spectra used for soil radiative property inputs were generated by the SOGM (average of 5 spectra) based on different input soil properties, which are shown in Fig. \ref{fig:3spectra}a. For Fig. \ref{fig:soilimages}a, the soil model input properties were determined based on \textit{Terra Preta} soil as described in \citep{neina2023terra} and \citep{eden1984terra}. \textit{Terra Preta} soil is nearly black in color, which is reproduced in the synthetic image (Fig. \ref{fig:soilimages}a). The soil model input properties for Fig. \ref{fig:soilimages}b were determined based on soil properties near Davis, CA, United States recorded in the Soil Survey Geographic (SSURGO) Database. More specific details regarding input soil parameters are given in the figure captions corresponding to each image.

\begin{figure}
\includegraphics[width=\textwidth]{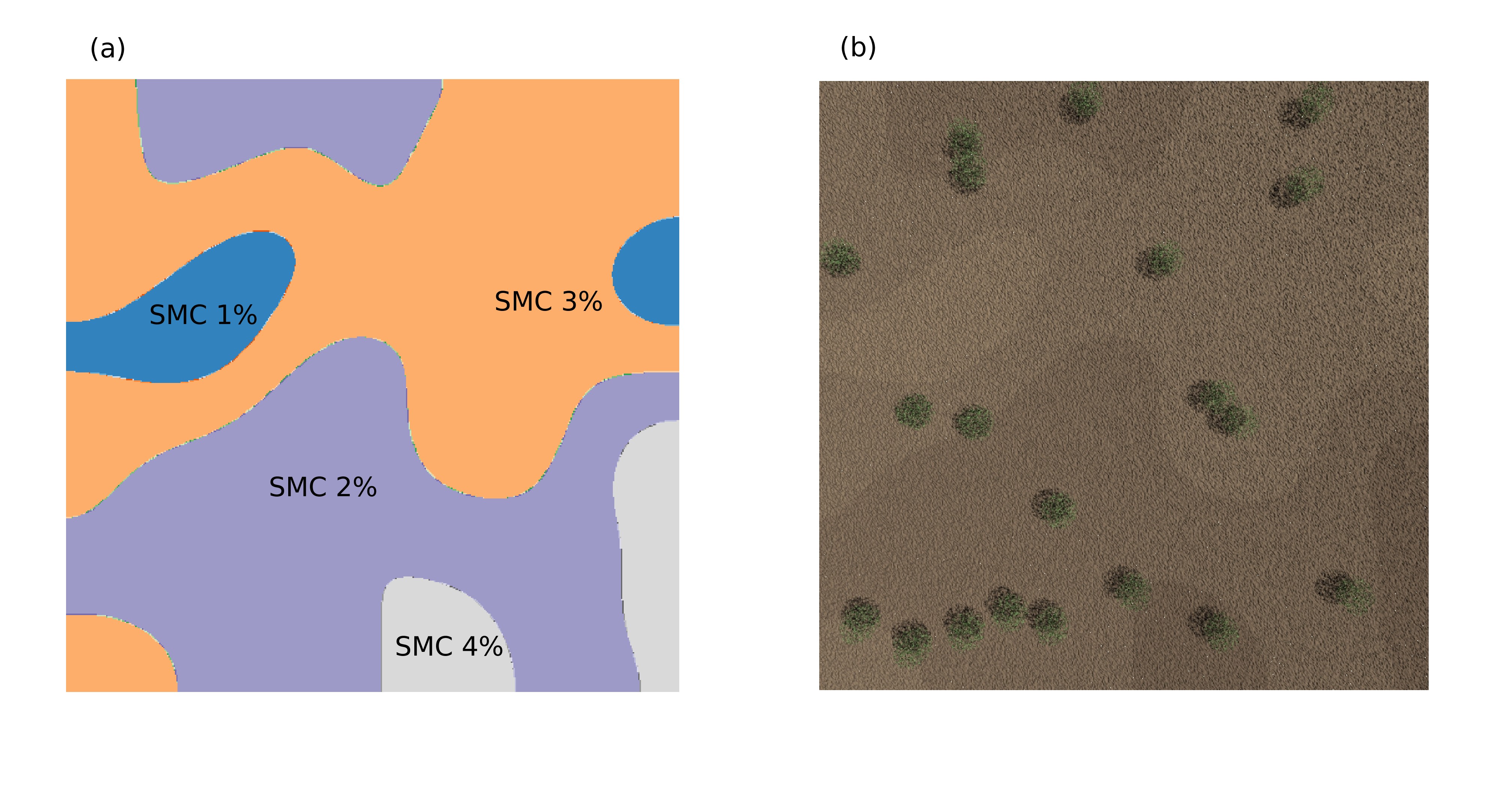}
\caption{(a) Soil SMC distribution map. (b) Synthetic soil image based on soil spectra affected by corresponding SMCs.}
\label{fig:soilmap}
\end{figure}

Figure \ref{fig:soilmap} displays a synthetic image featuring one type of soil with different SMCs, along with its corresponding SMCs distribution map. Since users can specify the soil properties, the 3D ray-tracing model labels the soil image at the pixel scale, which can be used as training data for machine learning models.

\begin{figure}
\includegraphics[width=\textwidth]{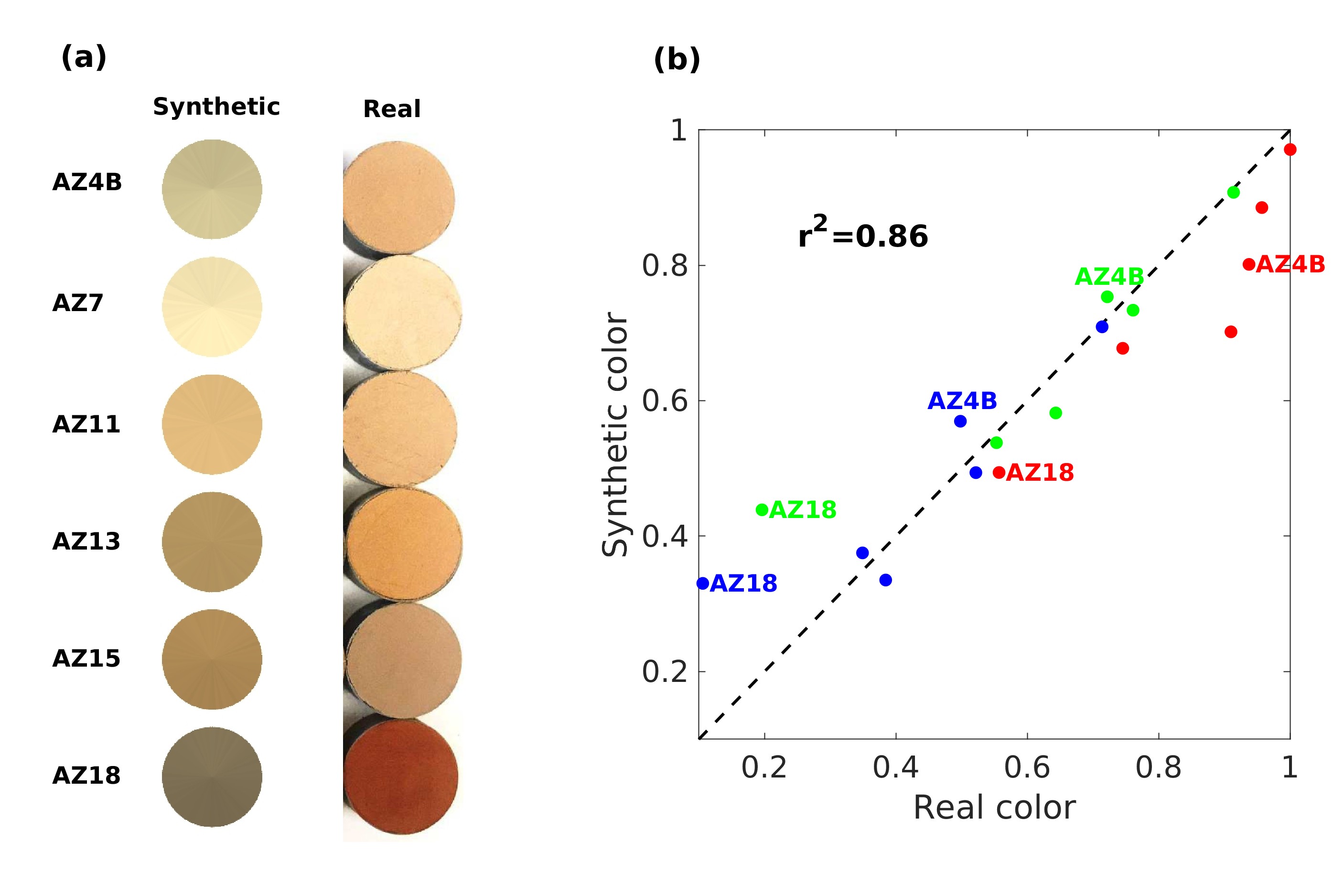}
\caption{Comparison of real and synthetically generated images of soil samples with different properties. (a) Real soil images from \citep{sadeghi2018particle} (right) compared against synthetic images generated using Helios with soil reflectance spectra generated from the SOGM (left). The synthetic image was generated based on properties provided in the original paper \citep{sadeghi2018particle} can be found in Table \ref{tab:paperprop}. Sample IDs are also in accordance with those provided in the paper. (b) Real and synthetic soil RGB values represent the mean pixel values of corresponding soil sample regions, scaled from 0$\sim$255 to 0$\sim$1. Data points for samples AZ4B and AZ18 are labeled for comparison against the images in (a).}
\label{fig:paperimage}
\end{figure}

Figure \ref{fig:paperimage}a shows synthetic soil images based on properties reported by \citep{sadeghi2018particle} and corresponding real soil images. As is normally the case, the lighting type and camera model used to create the real soil photographs was not reported. We therefore assumed that the soil samples were illuminated by 4 Cree XLamp XHP70.2 LED light sources, and that a Basler ace acA2500-20gc RGB camera (Basler, Ahrensburg, Germany) was used to capture the image. The generated soil reflectance spectra and simulated camera spectra responses are shown in Fig. \ref{fig:paperspec}. The synthetic red color values of soil sample AZ4B and AZ11 are lower than their actual red color values, which is a primary factor in the visual differences observed between their real and synthetic soil images. This discrepancy may be due to the absence of specific light and camera settings, particularly the bias of the red camera spectral response and blue peak in the LED light source spectrum.  Additionally, there is a noticeable difference between the real and synthetic green and blue color values for soil sample AZ18 (Fig. \ref{fig:paperimage}), which is likely caused by the limited soil properties provided by \citep{sadeghi2018particle}. Despite these limitations, the synthetic color values derived from the simulated soil images still exhibit a high $r^2$ of 0.86 with actual color values. This demonstrates the robustness and accuracy of the spectra generated by SOGM and images simulated by the Helios ray-tracing model.

\section{Discussion}

The present SOGM can generate soil spectra based on a wide range of soil properties such as particle size, organic matter content, organic carbon content, total nitrogen content, iron content, cation-exchange capacity, pH value, bulk density, and SMCs. 
The biggest contribution of the present work is to build a model that can encode most of the soil reflectance spectra-property data available online through a text-based generative model, and remove the barrier due to data non-uniformity such between mass-based and volumetric concentrations. Many radiative transfer based models that can simulate soil reflectance spectra in the VIS-NIR region rely on only one or a few properties, which mainly focus on soil moisture \citep{marmit,marmit2,jacquemoud1992modeling,bach1994modelling}, particle size \citep{sadeghi2018particle,wu2023semi}, and organic matter \citep{wu2023semi}. Compared to these models, the SOGM offers greater flexibility by accepting any combination of soil properties as inputs. When the input soil property set is incomplete, the model generates a spectrum that is reasonable within the constraints of available parameters. As a more complete property set is input to the model, uncertainly in the model is reduced and the generated spectrum becomes more constrained. 

The SOGM was tested on three datasets that were not included in the training process, which produced good results. Although the performance of wet soil spectra simulations was not as good as the multilayer radiative transfer model of soil reflectance (MARMIT) \citep{marmit} and MARMIT2 \citep{marmit2}, the present model can be directly applied on new dry soil spectra without extracting specific model parameters. By combining with the 3D ray-tracing model in Helios, soil images can also be generated based on available soil properties to generate soil that appears reasonable based on these properties.

Apart from soil research, the current model can enable investigation of a variety of radiation-dependent soil-plant interactions by combining the SOGM with other radiation modeling tools such as Helios \citep{HELIOS} and PROSAIL \citep{prosail}. The Helios 3D modeling software has sub-models for plant biophysical processes, thus adding the SOGM can allow for generalized specification of soil optical properties in order to improve representation of plant-soil interactions. Furthermore, the 3D ray-tracing model in Helios supports automatic image annotation (allowing the assignment of traits to individual pixels) that incorporates multiple plant traits such as plant height, leaf chemical concentrations, plant or leaf ID for object detection, etc. Adding the SOGM can further extend the capabilities to soil property labelling. Although not explicitly investigated in present study, the synthetic images can serve as inputs to machine learning models designed for remote and proximal sensing studies. While the example applications focused on RGB imagery datasets, another important strength of the proposed modeling framework is that it can simulate other sensor modalities such as multispectral imagery. The reflectance spectra can be integrated across arbitrary wavelength bands or camera spectral response curves. The integration of the current SOGM with the PROSAIL model \citep{prosail} is straightforward, with details provided in \ref{AppA}. As the soil reflectance generation is based on a wide range of user-specified properties, this combination may enable many potential remote sensing studies linking soil properties to complex surface radiative properties. 

The SOGM has some limitations that should be considered during its application. Since the model is data-driven, inaccuracies in the training data can translate into inaccuracy in model predictions. Most notably, certain minor soil properties with low concentration such as nickel might adversely affect spectra generation due to their high propensity for measurement errors and small influence on soil spectra (Table \ref{tab:data4test_result}). This can be mitigated by removing these properties from the list of inputs when possible. Additionally, there is a lack of wet soil spectra available in public spectral repositories, leading to relatively high prediction errors for the wet soil spectra model. Increasing the availability of open spectral datasets could help mitigate these limitations in the future.




\section{Conclusion}

In this research, we introduced the SOGM, an new model for generating soil VIS-NIR reflectance spectra based on incomplete input physical properties with variable formats. The model's capability to encode a vast array of soil spectra-property data available online through a text-based generative approach sets it apart from existing models, overcomes the limitations of data incompatibility barriers, and offers the flexibility to process various combinations of soil attributes, including soil moisture, without any additional model parameters. 

This work demonstrated the possibility of using a data-driven approach for modeling soil reflectance based on diverse and incomplete datasets. 
Through testing of the SOGM on new datasets not included in model training, this work demonstrated that the model can generate reasonable soil reflectance spectra based on available property inputs, which can be improved and constrained as the input parameter set increases (although this may not always be the case depending on the accuracy of the input property measurements). Results suggested that the most widely used soil properties, including particle composition (clay/sand/silt), bulk density, organic carbon, organic matter, and total nitrogen contents tended to improve soil reflectance predictions when available. Properties such as nickel content tended to decrease model performance due to their relatively low concentrations and elevated propensity for measurement error.

The integration of the SOGM with a radiation-based image simulator such as Helios, LargE-Scale remote sensing data and image simulation framework (LESS) \citep{LESS}, and the Discrete Anisotropic Radiative Transfer (DART) Lux model \citep{DARTlux1,DARTlux2} enables the generation of realistic soil images based on diverse soil properties. This combination paves the way for comprehensive studies in soil-plant interactions. The potential of our model in supporting remote sensing studies, particularly when integrated with the soil-plant radiative models such as PROSAIL model, is noteworthy. Despite some limitations, such as the impact of certain soil properties on spectra generation and the scarcity of wet soil spectra data, the SOGM represents a substantial step forward in soil spectra simulation.

This study not only contributes to the advancement of soil spectra generation but  also opens new ways for future exploration and innovation in the field of ecosystem and agriculture.

\section*{Declaration of Competing Interest}
The authors declare that they have no known competing financial interests or personal relationships that could have appeared to influence the work reported in this paper.

\section*{Acknowledgements}
This work was supported, in whole or in part, by the Bill \& Melinda Gates Foundation INV-0028630. Under the grant conditions of the Foundation, a Creative Commons Attribution 4.0 Generic License has already been assigned to the Author Accepted Manuscript version that might arise from this submission.

\appendix
\gdef\thesection{Appendix \Alph{section}}

\section{Integration with the PROSAIL model}\label{AppA}

\begin{figure}
\includegraphics[width=\textwidth]{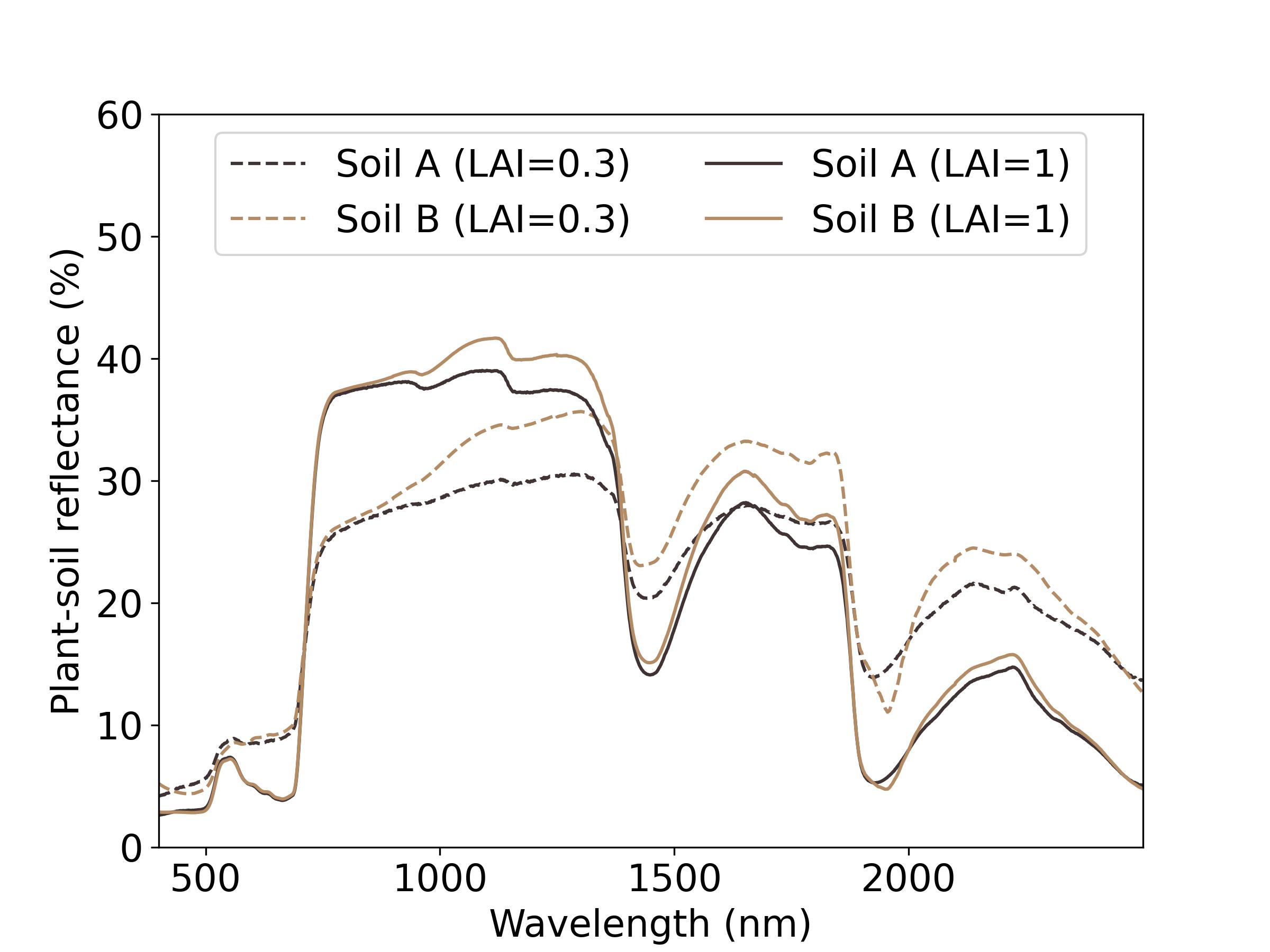}
\caption{Soil A and B use the input properties corresponding to those in Fig. \ref{fig:3spectra}a and Fig. \ref{fig:3spectra}b, respectively. Other PROSAIL inputs include: the number of elementary layers set to 1.5, chlorophyll concentration at 40 $\mu$g/cm, carotenoid concentration at 8 $\mu$g/cm, equivalent water thickness at 0.01 g/cm, dry matter content at 0.009 g/cm, and Solar zenith angle at 30\degree. Two LAI value, 0.3 and 1, are employed for the simulations of two different soil-plant surface spectra.}
\label{fig:prosailspec}
\end{figure}

The PROSAIL model is a fusion of the PROSPECT-based models \citep{PROSPECT, PROSPECTD, PROSPECT45} and SAIL-based models \citep{sail, geosail}. The 4SAIL model can be depicted mathematically as:

\begin{equation}
\mathbf{R_{surface}} = \mathrm{4SAIL}(\mathbf{R_{leaf}}, \mathbf{T_{leaf}}, \text{LAI}, \text{LIDF}, S_L, \theta{_s},\theta{_v},\mathbf{R_{soil}}),
\end{equation}

\noindent where $\mathbf{R_{surface}}$ is the surface reflectance spectra, $\mathbf{R_{leaf}}$ and $\mathbf{T_{leaf}}$ are leaf reflectance and transmittance spectra, which are obtained using the PROSPECT-based models by specifying leaf chemical properties such as chlorophyll concentration, carotenoid concentration, water concentration etc., $\text{LIDF}$ represents the leaf inclination distribution function containing one or two parameters, $S_L$ is the hot spot parameter, $\theta{_s}$ and $\theta{_v}$ denote solar and viewing zenith angle, respectively, and $\mathbf{R}_{soil}$ is the soil Lambertian reflectance spectra, which can be easily generated by our SOGM. Figure \ref{fig:prosailspec} shows example soil-plant spectra simulation using the integration of SOGM and PROSAIL model.

\section{Camera spectral responses}\label{AppB}

\begin{figure}
\includegraphics[width=\textwidth]{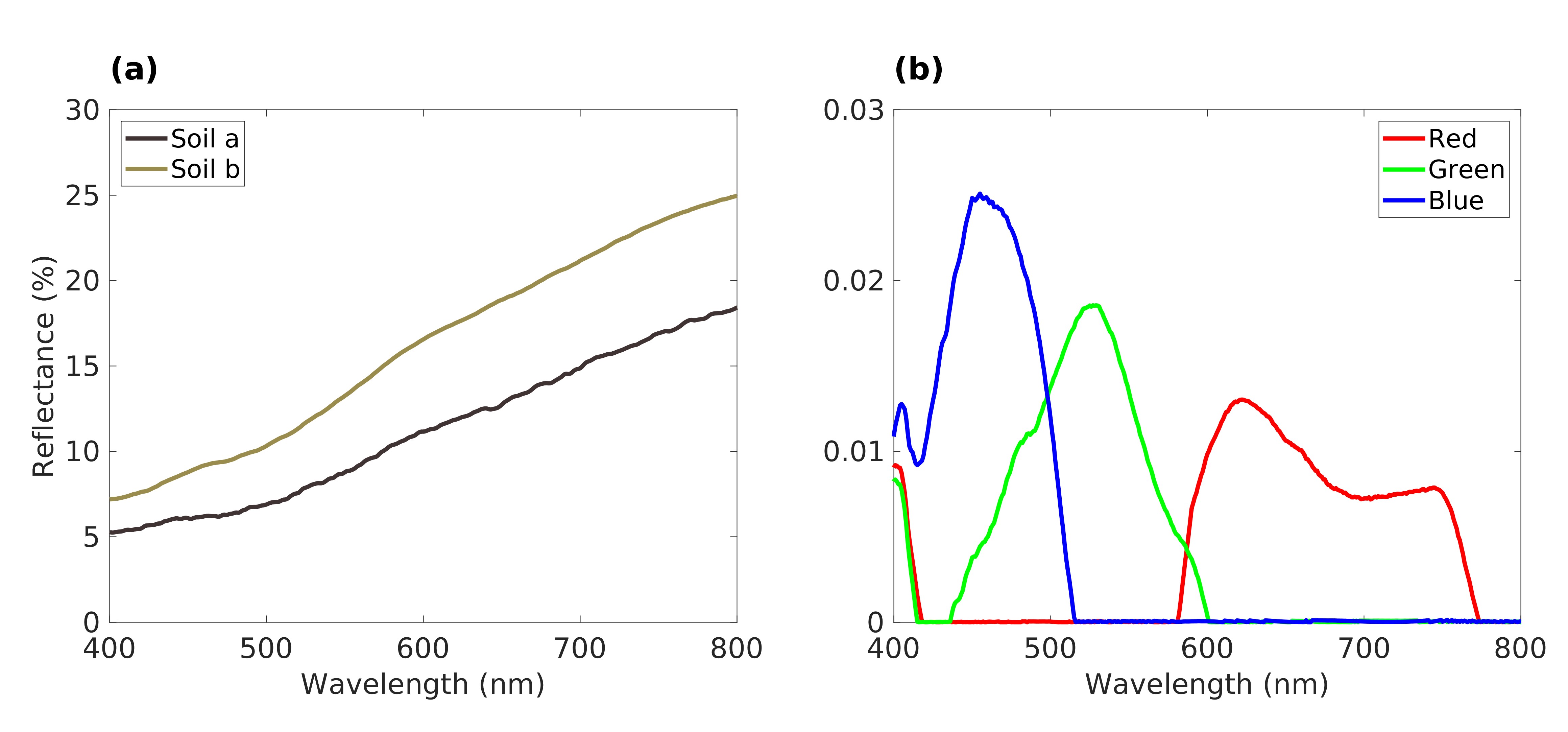}
\caption{(a) Synthetic soil spectra in visible region (400-800 nm) for generating synthetic soil image Fig. \ref{fig:soilimages}. (b) Calibrated Nikon B500 camera spectral responses. }
\label{fig:3spectra}
\end{figure}

\begin{figure}
\includegraphics[width=\textwidth]{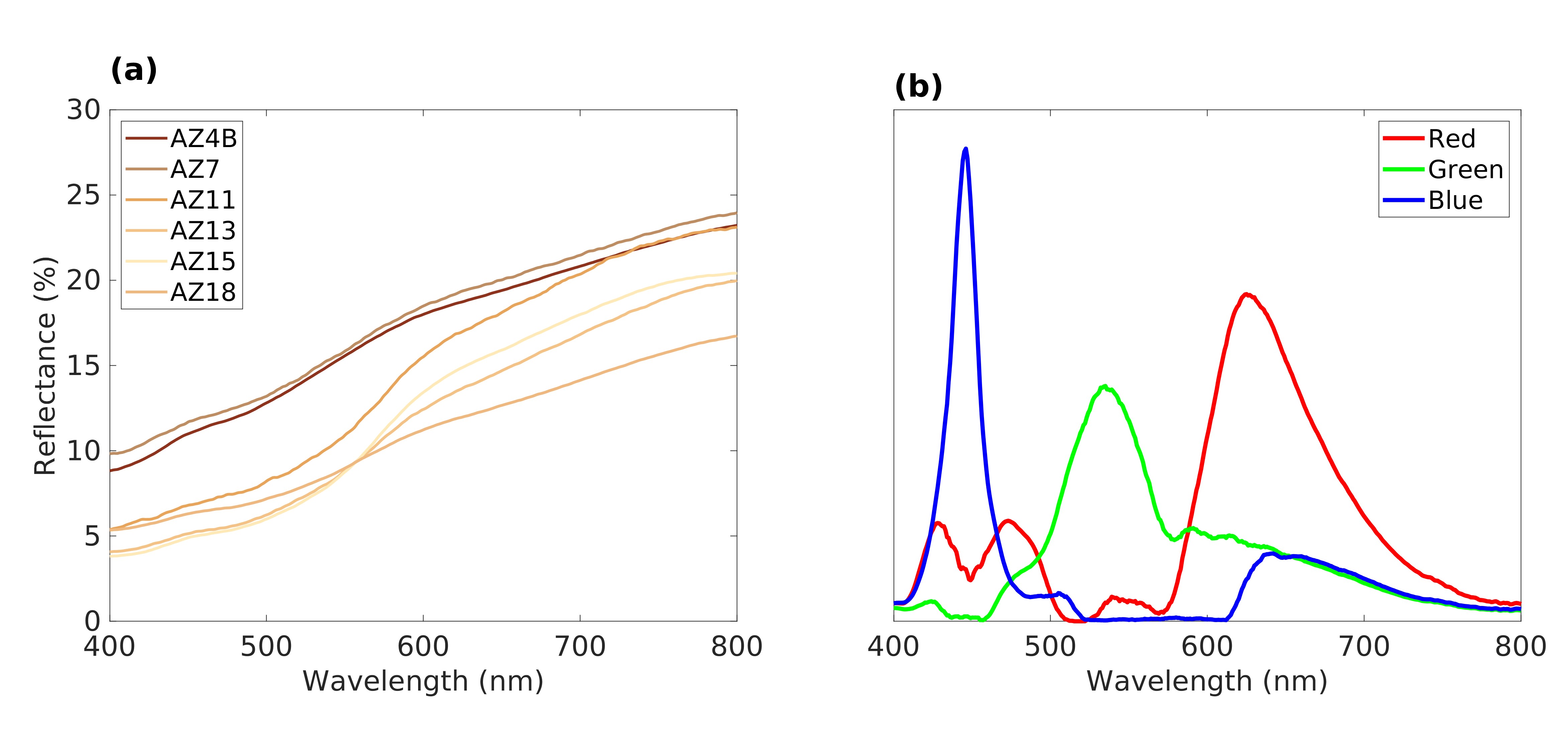}
\caption{(a) Synthetic soil spectra based on properties provided by \citep{sadeghi2018particle}'s literature in visible region (400-800 nm) for generating synthetic soil image Fig. \ref{fig:paperimage}a. (b) Calibrated Basler ace acA2500-20gc RGB camera spectral responses.}
\label{fig:paperspec}
\end{figure}

Figure \ref{fig:3spectra}a and \ref{fig:paperspec}a display soil reflectance spectra generated by SOGM, which were used to simulate soil-plant images in Fig. \ref{fig:soilimages} and \ref{fig:paperimage}, respectively. Given that the target images are in RGB bands, only the spectra within the visible region 400-800 nm are presented.

Figure \ref{fig:3spectra}b and \ref{fig:paperspec}b present the calibrated Nikon B500 (Nikon, Minato City, Tokyo, Japan) and Basler ace acA2500-20gc RGB (Basler, Ahrensburg, Germany) camera spectral responses used for generating Fig.\ref{fig:soilimages} and Fig. \ref{fig:paperimage}a, respectively. Table \ref{tab:paperprop} showcases the soil properties provided by \citep{sadeghi2018particle}'s paper used for simulating corresponding soil spectra.

\begin{table}[ht]
\centering
\caption{Soil Properties provided in \citep{sadeghi2018particle}'s paper.}
\label{tab:soil_properties}
\begin{tabular}{llllll}
\hline
Soil Code & Bulk densities (g/cm$^3$) & Sand (\%) & Silt (\%) & Clay (\%) & Organic Matter (\%) \\
\hline
AZ4B  & 1.528 & 80.4 & 14.2 & 5.4  & 2.3 \\
AZ7   & 1.313 & 58.5 & 32.0 & 9.5  & 1.1 \\
AZ11  & 1.398 & 38.5 & 40.1 & 21.4 & 1.7 \\
AZ13  & 1.256 & 58.1 & 15.4 & 26.5 & 2.8 \\
AZ15  & 1.025 & 3.6  & 73.4 & 23.0 & 3.4 \\
AZ18  & 1.161 & 29.1 & 18.7 & 52.2 & 4.0 \\
\hline
\end{tabular}
\label{tab:paperprop}
\end{table}

The Nikon B500 camera spectral response was calibrated based on a DKC-Pro color board image (DGK Color Tools, Boston, Massachusetts, USA) under real sun. The Basler ace acA2500-20gc RGB camera spectral response was calibrated based on a SpyderCHECKR 24 color board image (Datacolor, Risch-Rotkreuz, Switzerland) illuminated by 4 Cree XLamp XHP70.2 LED light sources.


\bibliographystyle{unsrt}
\bibliography{Ref}

\end{document}